\documentclass[runningheads]{llncs}
\usepackage{graphicx}
\usepackage{caption} 
\usepackage{subcaption}
\usepackage{amsmath}
\usepackage{multirow}
\usepackage{xcolor,soul}
\begin{document}
\title{Logo Generation Using Regional Features: A Faster R-CNN Approach to Generative Adversarial Networks\thanks{accepted in EAI ArtsIT 2021}}
%
\titlerunning{Logo Generation}
%
\author{Aram Ter-Sarkisov \and Eduardo Alonso}
\authorrunning{}
%
\institute{CitAI Research Center \\Department of Computer Science \\City, University of London\\ \email{alex.ter-sarkisov@city.ac.uk}}  
%
\maketitle              
\vspace{-10pt}
\begin{abstract}
In this paper we introduce Local Logo Generative Adversarial Network (LL-GAN) that uses regional features extracted from Faster R-CNN for logo generation. We demonstrate the strength of this approach by training the framework on a small style-rich dataset of real heavy metal logos to generate new ones. LL-GAN achieves Inception Score of 5.29 and Frechet Inception Distance of 223.94, improving on state-of-the-art models StyleGAN2 and Self-Attention GAN.  

\keywords{Deep Learning  \and Generative Adversarial Networks \and Logo Generation.}
\end{abstract}
\section{Introduction}
\label{intro}
Generative Adversarial Networks (GANs) were first introduced in \cite{goodfellow2014generative}. They have gained a wide recognition in the Artificial Intelligence community due to their ability to approximate the distribution of real data by generating fake data. Recent advances include Progressive-Growing GANs, StyleGAN and StyleGAN2 that learn styles at different resolutions\cite{karras2017progressive,karras2019style,karras2020analyzing}, Self-Attention GANs (SAGANs) that learn the connections between different spatial locations\cite{zhang2019self}, CycleGANs and Pix2Pix GANs for unpaired style transfer\cite{zhu2017unpaired,isola2017image} and Wasserstein loss function\cite{arjovsky2017wasserstein}.\\ 
\linebreak
Faster R-CNN and Mask R-CNN\cite{ren2015faster,girshick2014rich,he2017mask} are state-of-the-art open-source deep learning algorithms for object detection and instance segmentation that work in multiple stages, unlike single-shot models like YOLO\cite{redmon2016you}.\\ 
\linebreak
Faster R-CNN first predicts regions containing objects based on overlaps (Intersect over Union, IoU) between fixed-size rectangles known as anchors and ground truth bounding boxes using Region Proposal Network (RPN). Then, it pools features from these areas by cropping and resizing corresponding areas in features maps. This is done using Region of Interest Pooling (RoIPool) to construct fixed-size Regions of Interest (RoIs) containing rescaled regional features for each object (later replaced by more accurate Region of Interest Align, RoIAlign\cite{he2017mask}). These local features are fed through fully connected (\texttt{fc}) layers to independently predict the object classes and refine bounding box prediction. In addition to this, Mask R-CNN segments objects' masks.\\    
\linebreak
One of the new and challenging areas in GANs and neural style transfer is the creation of logos and fonts. This area includes style and shape transfer between fonts\cite{azadi2018multi,gao2019artistic}, logo synthesis\cite{sage2018logo,oeldorf2019loganv2,mino2018logan}, transfer of style to font\cite{atarsaikhan2018contained} and font generation\cite{hayashi2019glyphgan}. A specific challenge in this area is disentaglement of content and style learning, often done through training of two different encoders and feature concatenation, as in \cite{gao2019artistic}, and separation of transfer of shape and texture (ornamentation), done through pretraining of the shape model and ornamentation model that takes the shapes and adds ornamentation\cite{azadi2018multi}. Logo synthesis (style transfer), as in \cite{oeldorf2019loganv2,mino2018logan,sage2018logo}, also uses conditional input (random vector + sparse vector for the class). \\    
\linebreak
We address the shortcomings of the state-of-the-art models, such as the size of the output, which in most cases is limited to 64x64 pixels. This size is sufficient for separate characters/glyphs or small logos, as readability does not suffer. For larger logos or words, model output must be upsampled. Another limitation we address is the size of the training data: we leverage Faster R-CNN's capacity to sample a batch of regional features in a single image to overcome the need for a large dataset.\\    
\linebreak
In this paper we present a GAN model for generating logos of heavy metal bands. To the best of our knowledge, it would be the first GAN study that is focused on the generation of band logos. With respect to specifically heavy metal logos, recently, there were two related publications: in \cite{ter2020network} style transfer model based on \cite{gatys2016image} was used to fuse the style of heavy metal bands logos, e.g. Megadeth and the content of corporate logos, e.g. Microsoft. In \cite{rijken2021illegible} the styling of heavy metal logos and its association with genre and readability are investigated.\\
\linebreak
Measured by Frechet inception distance\cite{heusel2017gans}, Inception score\cite{salimans2016improved} and detection accuracy, the presented model confidently outperforms the state-of-the-art StyleGAN2 and SAGAN frameworks. Our contribution consists of the following: 
\begin{itemize}
    \item Local Logo GAN (LL-GAN) framework: training the Generator by comparing regional features extracted from the fake and real data using RoIAlign module in Faster R-CNN. Since loss is computed only on regional features, the Generator's parameters receive updates only from the region containing the logo in the real data. This model augments the baseline GAN framework, serving as an additional source of gradients for the Generator's parameters. Ground truth bounding box is used to determine positive RoIs in the fake image, therefore the Generator learns to output spatially-aware logos. A number of RoIs is sampled from each image using RPN and RoIAlign modules, which compensates for the sparsity of the data,\\
    \item Logo generator. The model is capable of generating style-rich heavy metal logos consisting of glyph-like structures that closely resemble real-life band logos without suffering from the mode collapse. This includes an augmentation of the DCGAN's model architecture\cite{radford2015unsupervised} that allows for creation of large images (282$\times$282),\\    
    \item Style-rich metal band logos dataset. Images with heavy metal band logos were scraped from the internet and labelled at text level (bounding box around the band's logo). Each image contains a single-word logo, with a simple background (e.g. black or white) across 10 bands selected for the style of the logo. The dataset consists of 923 images and an equal number of bounding box coordinates of the logo.
\end{itemize}
\begin{figure*}
\vspace{-10pt}
    \centering
    \includegraphics[width=1.0\linewidth,height=0.5\linewidth]{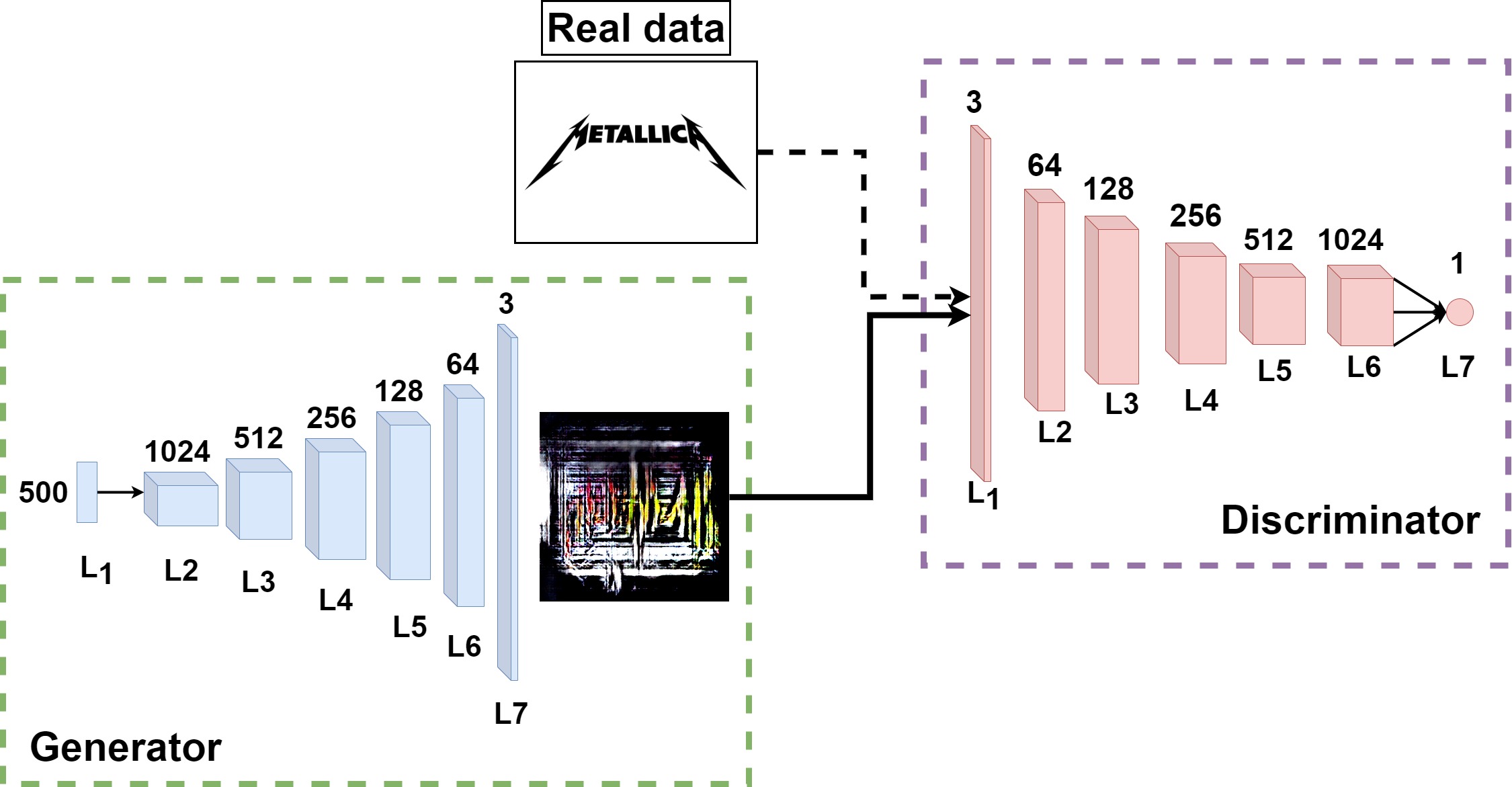}
    \caption{DCGAN+ framework. Details of the architecture of both models is presented in Table \ref{tab:dcgan_plus}. Values in each module in the number of feature maps in the Convolution (Discriminator) or Transposed Convolution (Generator) models. Normal arrows: features and fake data, broken arrow: real data.}
    \label{fig:dcgan_alex}
    \vspace{-25pt}
\end{figure*}
\section{Our Approach}
\label{sec:our_method}
Model sizes and structures are compared in Table \ref{tab:model_sizes}. 
\subsection{DCGAN+ framework} 
DCGAN+ is an augmentation of the DCGAN architecture\cite{radford2015unsupervised} that enables generation of larger images in a single shot. The main idea behind the architecture is the selection of the right rate of upsampling and downsampling of feature maps in each model (kernel size, stride, padding). Figure \ref{fig:dcgan_alex} and Table \ref{tab:dcgan_plus} provide a summary of the models' architectures. This solution successfully addresses the problem of the size of the generated logo, as we increase it from at most 64$\times$64, as in \cite{sage2018logo} to 282$\times$282.\\
\begin{table*}
\centering
\vspace{-30pt}
    \caption{Comparison of sizes of the frameworks. G: generator, D: discriminator, F: Faster R-CNN.}
    \vspace{+5pt}
\begin{tabular}{lll}
Framework & Number of Parameters & Structure of the framework\\
\hline
    DCGAN+     &43.83M + 3.93M & G + D \\
    LL-GAN     &43.83M + 3.93M + 41.43M & G + D + F      \\
\hline
    StyleGAN2 \cite{karras2020analyzing}& 84.69M (Total) & G + D\\
    StyleGAN2 w/attention \cite{karras2020analyzing}& 85.87M (Total) &G + D \\
    \hline
    SAGAN \cite{zhang2019self} &8.1M + 4.92M & G + D\\
    \hline
    DCGAN \cite{radford2015unsupervised}& 3.5M + 2.7M &G +D \\
    Faster R-CNN \cite{he2016deep}& 41.80M& F\\
  \end{tabular}
    \label{tab:model_sizes}
\end{table*}
\begin{table*}
    \centering
    \vspace{-35pt}
    \caption{DCGAN+ framework. G: Generator, D: Discriminator} 
    \vspace{+5pt}
    \begin{tabular}{l|lllll}
         Model&Block&Depth&Kernel&Stride&Pad\\
         \hline
         \multirow{7}{*}{G}&$L_1$(\text{Input})&500&0&0&1\\
         &$L_2$&1024&8&2&0\\
         &$L_3$&512&4&2&0\\
         &$L_4$&256&4&2&1\\
         &$L_5$&128&4&2&1\\
         &$L_6$&64&2&2&1\\
         &$L_7$(\texttt{tanh})&3&2&2&1\\
         \hline
         \multirow{7}{*}{D}&$L_1$(Input)&3&&&\\
         &$L_2$&64&4&2&1\\
         &$L_3$&123&3&2&1\\
         &$L_4$&256&3&2&1\\
         &$L_5$&512&3&2&1\\
         &$L_6$&1024&3&2&1\\
         &$L_7$(\texttt{fc})&1&-&-&-\\
    \end{tabular}
    \vspace{-15pt}
    \label{tab:dcgan_plus}
\end{table*}   
 \begin{figure*}
    \centering
    \includegraphics[width = .8\linewidth, height = 0.4\linewidth]{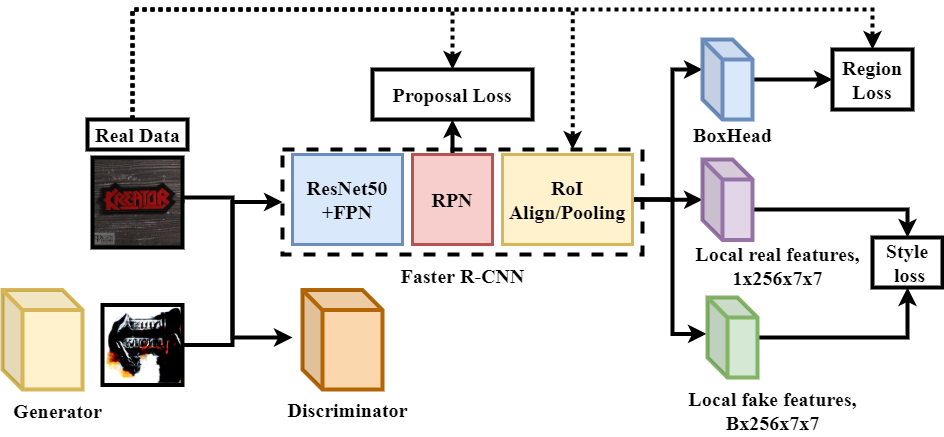}
    \caption{LL-GAN framework. Normal arrows: features, dotted arrows: box coordinates, broken line box: Faster R-CNN.}
    \label{fig:frcnn_a}
    \end{figure*}
\subsection{LL-GAN framework} 
Overall framework is presented in Figure \ref{fig:frcnn_a}. Generator and Discriminator are the same as in DCGAN+.  One of the key contributions of this paper is the use of local features from the RoIAlign stage in Faster R-CNN to compute style loss. We use the ground truth bounding box around the band logo to extract one RoI from the real data, skipping the RPN stage. For the fake data, RPN predicts raw boxes passed on to RoIAlign that uses these predictions to extract RoI features and outputs $B$ positive predictions (i.e. RoI box predictions that have IoU with the ground truth box greater than a pre-defined threshold), each of fixed size $H \times W \times C$. Each RoI's height and width are hyperparameters, and depth $C$ is determined by the depth of the FPN feature map, see \cite{lin2017feature}.\\ 
\linebreak
Feature loss is computed between $B$ positive RoIs from the fake and the single RoI from the real data (ground truth region). The number of RoIs varies from image to image, but on the average grows as the fake data increasingly resembles the real data.\\
\linebreak
Each of $C$ feature maps extracted from the real data is vectorized, i.e. an $i^{th}$ feature map is converted into a vector with $H \cdot W = HW$ elements which we refer to as $\mathcal{F}^{r}_{i}$. Dot-product is computed between each $(i,j)$ pair of vectorized feature maps to obtain matrix $\mathcal{G}^{r}$ with dimensionality $C \times C$ (i.e. each $(i,j)$ element in $\mathcal{G}^{r}$ is a dot product of the vectors $\mathcal{F}^{r}_{i}$ and $\mathcal{F}^{r}_{j}$), see Equation \ref{eq:gram_matrix_real_data}.
\begin{equation}
    \mathcal{G}^{r}_{i,j} = \mathcal{F}^{r}_{i} \otimes \mathcal{F}^{r}_{j}
    \label{eq:gram_matrix_real_data}  
\end{equation}
For each $k^{th}$ RoI extracted from the fake data, we also compute Gram matrix $\mathcal{G}^{k, f}$, Equation \ref{eq:gram_matrix_generated_data}, where $\mathcal{F}^{k, f}_{i}$ is an $i^{th}$ vectorized feature map in the $k^{th}$ RoI. Therefore $\mathcal{G}^{k, f}_{i,j}$ is the dot-product between each $(i,j)$ pair of vectorized feature maps in $k^{th}$ RoI, $\mathcal{F}^{k, f}_{i} \otimes \mathcal{F}^{k, f}_{j}$.
\begin{equation}
    \mathcal{G}^{k, f}_{i,j} = \mathcal{F}^{k, f}_{i} \otimes \mathcal{F}^{k, f}_{j}
    \label{eq:gram_matrix_generated_data}  
\end{equation}
Equations \ref{eq:gram_matrix_real_data} and \ref{eq:gram_matrix_generated_data} compute correlation between regional features, which represents the style. The normalized style loss of $k^{th}$ RoI, $D_{k}$ is computed using $L_2$ distance between $\mathcal{G}^{r}$ and $\mathcal{G}^{k, f}$ elementwise, Equation \ref{eq:style_loss_fm}. Finally, we sum $B$ normalized RoI losses, Equation \ref{eq:style_loss}. 
\begin{align}
    D_k &= \frac{\sum_{i=1}^{C}\sum_{j=1}^{C}\big(\mathcal{G}_{i,j}^r - \mathcal{G}_{i,j}^{k, f}\big)^2}{(2\times H \times W)^2} \label{eq:style_loss_fm}\\
    L^{S} &=\frac{\sum_{k=1}^{B}D_k}{B}
    \label{eq:style_loss}
\end{align}
The main idea of computing style loss using Equations \ref{eq:gram_matrix_real_data} - \ref{eq:style_loss} is to train the Generator to evolve features that approximate the distribution of the real logos, and in the same region as in the real data. The first requirement (style) is satisfied by Equations \ref{eq:gram_matrix_real_data} and \ref{eq:gram_matrix_generated_data}, the second one (spatial awareness) by the RoIAlign functionality: by backpropagating loss extracted from a region in the fake data, Generator learns to evolve region-aware logos.  Total loss in this framework is computed using Equation \ref{eq:total_LL-GAN}.\\ 
\linebreak
\begin{align}
    L^{D} &= \mathbf{E}_{x \sim p(x)} \log D(x)  + \mathbf{E}_{z \sim p(z)}\log (1-D(G(z))) \label{eq:total_D}\\
    L^{G} &= \mathbf{E}_{z \sim p(z)}\log D(G(z)) \label{eq:total_G} \\
    L_{Total} &= L^{G} + L^D + L^{S} \label{eq:total_LL-GAN}
\end{align}
\linebreak
Equations \ref{eq:total_D} and \ref{eq:total_G} are the usual Discriminator and Generator losses, both computed using binary cross-entropy, for the real data $x$ and fake data $z$, except that Generator loss maximizes the loss function instead of minimizing it, see Section \ref{sec:experiments} for details. $L^{S}$ is the style loss in Equation \ref{eq:style_loss}.
\begin{figure*}
  \centering
  \includegraphics[width=0.11\linewidth, height = .15 \linewidth]{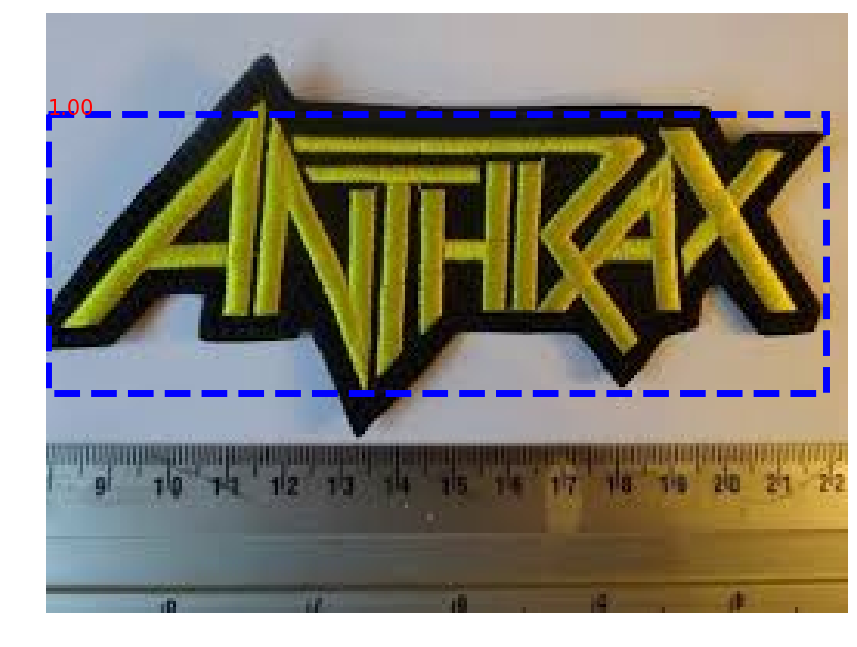}
  \includegraphics[width=0.11\linewidth]{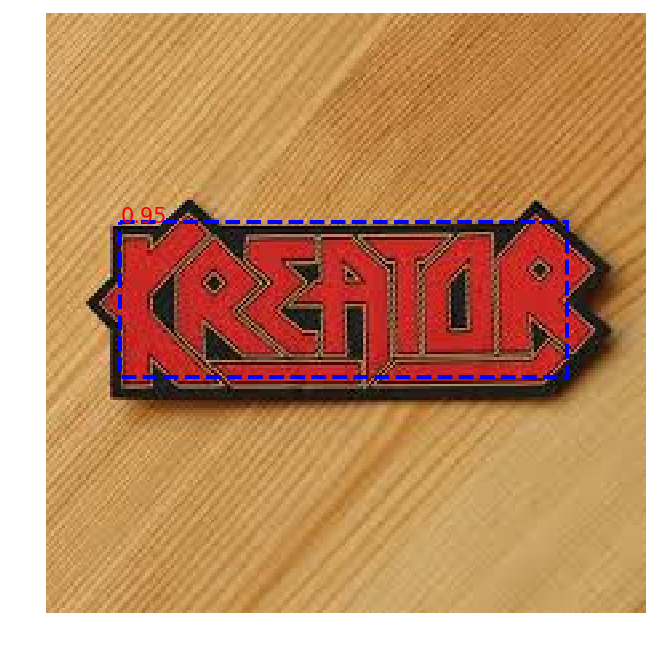}
  \includegraphics[width=0.11\linewidth]{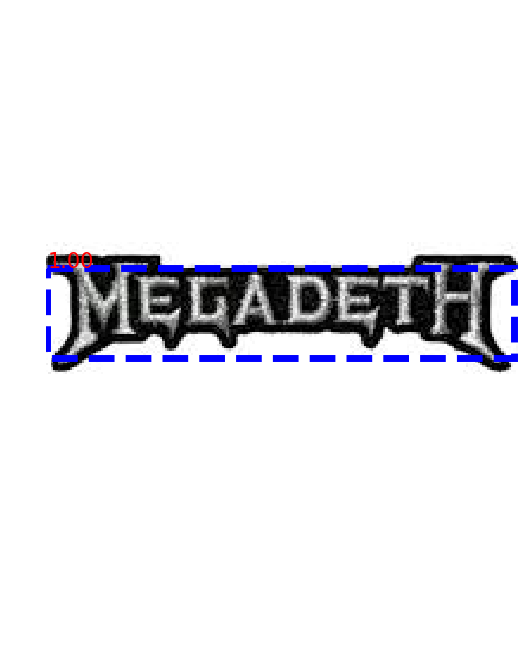}
  \includegraphics[width=0.11\linewidth]{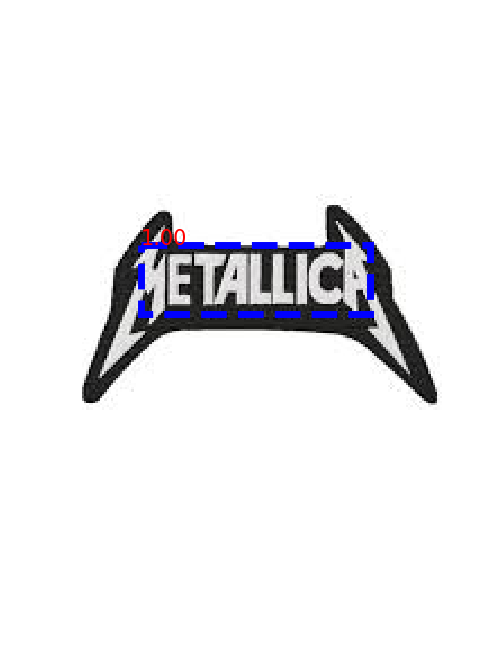}
  \includegraphics[width=0.11\linewidth]{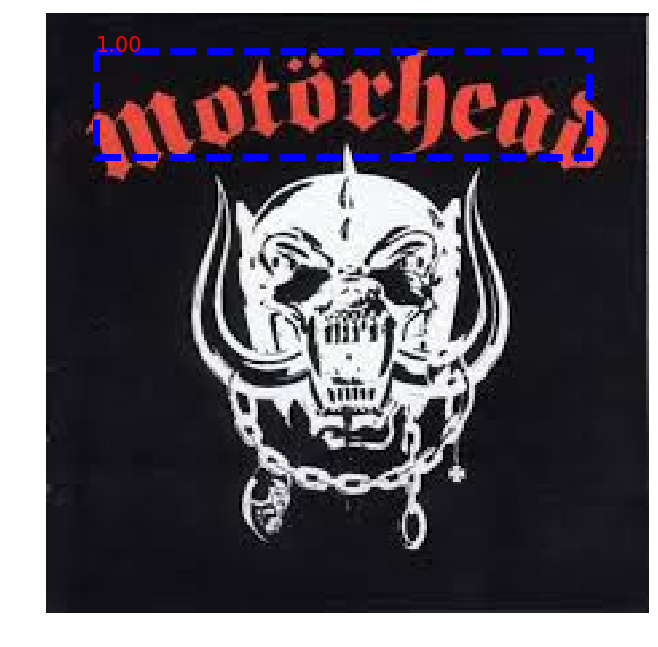}
  \includegraphics[width=0.11\linewidth]{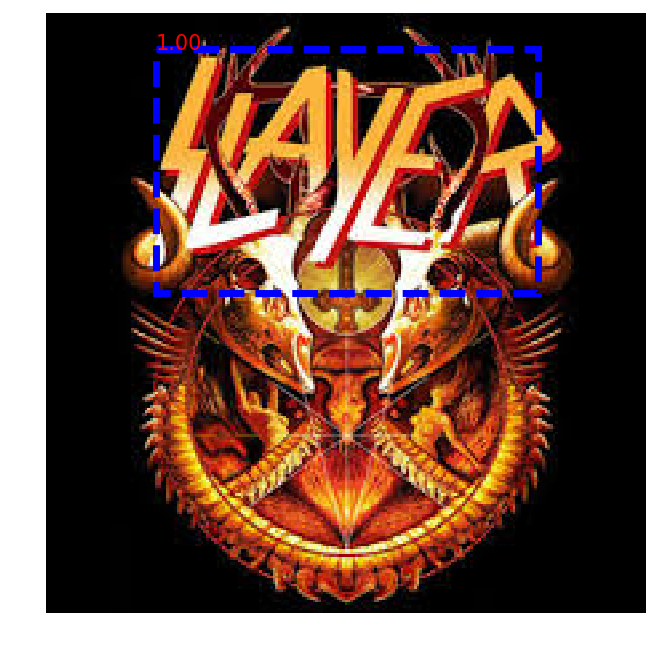}
  \includegraphics[width=0.1\linewidth]{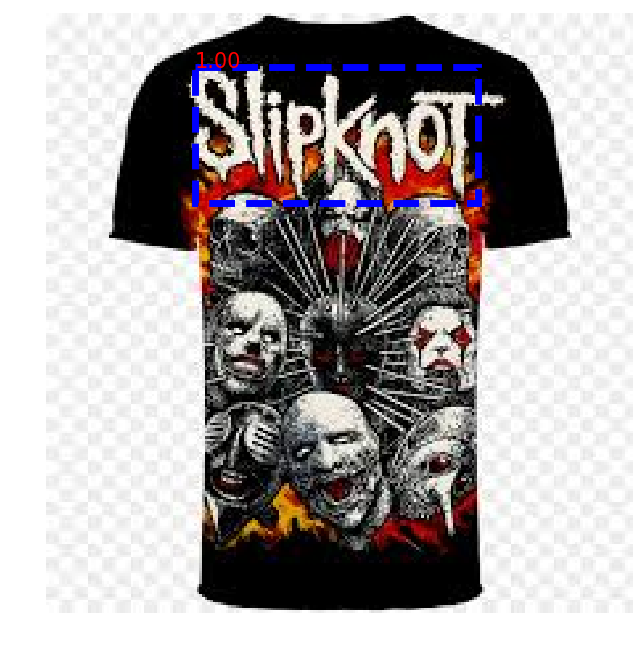}
  \includegraphics[width=0.11\linewidth]{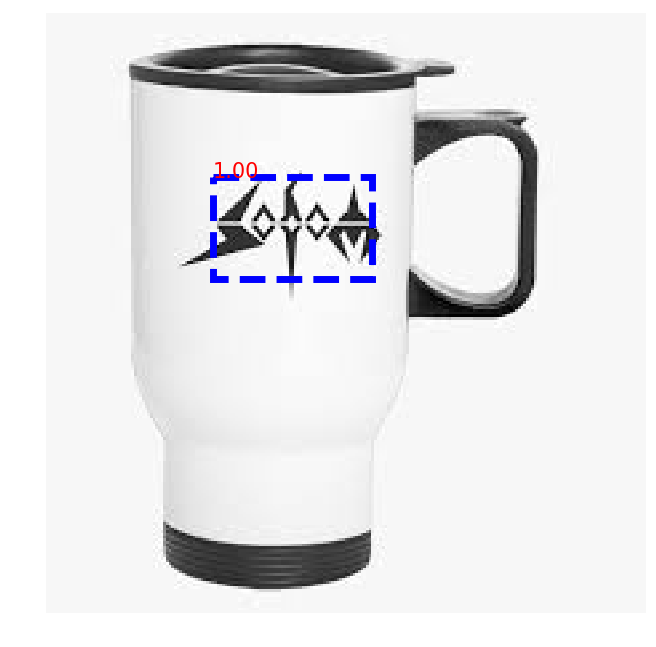}
   \caption{Examples of logos used in the training data overlaid with bounding box and score predictions by Faster R-CNN. Best viewed in color.}
   \label{fig:samples}
   \vspace{-15pt}
\end{figure*}
\section{Dataset construction and labeling}
\label{sec:data}
To train LL-GAN models, dataset must have labels consisting of bounding boxes around logos (one box per image). Therefore, dataset construction consists of three stages: first, we scrape the logos from the internet and manually labelled a small portion of it. Next, we train Faster R-CNN on a labelled text and logo ICDAR dataset, to predict boxes around words, and finetuned it to the labelled portion of the metal logo data. Finally, we use this model on the remaining scraped data to label each metal logo with the bounding box.
\subsection{Raw dataset} 
Our real dataset consists of 923 images of varying sizes. Each image contains a heavy metal band's logo, predominantly with a neutral (e.g. black or white) background. This was done in order to prevent the generator from learning background features and instead focus on the logo style and semantics. Ten bands were selected purely for the style of their logos: Anthrax, Kreator, Manowar, Megadeth, Metallica, Motorhead, Sepultura, Slayer, Slipknot, Sodom. The sizes of images vary between 50x50 and 512x1024 pixels, with the majority about 200x200. Examples with the overlaid bounding boxes are presented in Figure \ref{fig:samples}. This is a very challenging dataset, for two reasons: it is very small, and it is rich in style (specific styles of heavy metal logos/fonts) and weak in content, because each image contains only a single logo, there's a limited number of observations for each logo. As we explained in Section \ref{sec:our_method} and show in Section \ref{sec:experiments}, the ability of Faster R-CNN to learn and extract regional features from a single image addresses this challenge.   
\subsection{Faster R-CNN Logo Detector} 
To detect boxes around text in logos, we finetuned the out-of-the-box Faster R-CNN model from Torchvision v0.3.0 library with ResNet50 backbone feature extractor and FPN pretrained on MS COCO 2017 to ICDAR Focused Scene Text (ICDAR-FST2013), \cite{karatzas2013icdar} dataset that contains 223 images of street signs for 100 epochs. This model was trained to detect separate words in various contexts. Next, we fintetuned it for $500$ epochs to a portion of the metal logo dataset. The model predicts only two classes (object vs background) per RoI, and we capped the number of candidates in RPN stage at $1024$ and also used a slightly larger RPN anchor generator (5 anchor sizes between $16$ and $256$ and $5$ scales, between $0.25$ and $2$, a total of 25/location), learning rate of $1e-5$, regularization hyperparameter (weight decay) of $1e-2$ and Adam optimizer with $\beta_1=0.9, \beta_2=0.99$.  Other important hyperparameters (positive/negative box thresholds, RoI dimensions, RoI batch size, heads sizes) were the same as in the baseline Torchvision model. First, this model was used to label the rest of the metal logo data for experiments in Section \ref{sec:experiments}.Then, in Section \ref{sec:accuracy}, this model was used to detect logos produced by generators in all LL-GAN frameworks and to evaluate the accuracy of outputs of all generators and produce results in Table \ref{tab:accuracy_results}. 
\section{Experiments}
\label{sec:experiments}
\subsection{DCGAN+ framework} 
We trained both Generator and Discriminator in the DCGAN+ framework from scratch with a learning rate of $1e-4$ and weight regularization coefficient of $1e-3$ for both models using Adam optimizer \cite{kingma2014adam}, batch size of $128$ and binary cross-entropy loss for 1000 epochs. This took about 6 hours on a GPU with 8Gb VRAM. Following the recommendations in \cite{goodfellow2014generative} and Pytorch GAN tutorial, Discriminator is updated using real and fake data (1 iteration). Then, the fake data is relabelled as real and the Generator is updated by computing loss using real labels. This is done to avoid premature convergence.  
\subsection{LL-GAN framework} 
For LL-GAN we used the pretrained weights and the same architecture for the Generator and Discriminator from DCGAN+. Only Generator and Discriminator were trained, all Faster R-CNN weights trained in Section \ref{sec:data} remained frozen, since the logo detector model was specifically trained to detect single logos anywhere. Real and fake data is processed differently by the logo detector. From the real data, only single RoI regional features with dimensions $C \times H \times W$ is extracted and vectorized, Equation \ref{eq:gram_matrix_real_data}, using ground truth bounding box, hence RPN stage is skipped, and no gradients are computed. Fake data is fed forward through the whole framework (see Figure \ref{fig:frcnn_a}), RoI features are extracted and vectorized, Equation \ref{eq:gram_matrix_generated_data} for the loss, Equations \ref{eq:style_loss_fm}-\ref{eq:total_LL-GAN} and gradient computation. \\    
\linebreak
Also, RoI module, during processing of fake images, always appends the ground truth bounding box coordinates to the list of RoIs. The reason for that is that early in training, Generator cannot output high-quality logos, and therefore Faster R-CNN will not be able to find good RoIs anywhere in the fake data. As a result, the number of positive RoIs ($B$ in Equation \ref{eq:style_loss}) varied from image to image, but overall increased due to the improvement in the work of the Generator. In addition to the baseline LL-GAN framework that uses Equation \ref{eq:total_LL-GAN} loss function, we experimented with a number of tricks: 
\begin{itemize}
    \item In addition to style loss in Equation \ref{eq:style_loss}, we added detection loss from fake data. Ground truth bounding box coordinates were taken from the real logo that was used to train the Generator. This added two more loss functions: raw boxes in RPN and refined boxes in RoI,\\    
    \item Extend ground truth bounding boxes around logos to add more context when computing the Generator's loss. We experimented with different values and found 20 pixels in each direction the optimal number for the tradeoff between context and background noise.  \\    
    \item Compute $L_2$ loss between backbone features extracted from real and fake data, similar to content loss in neural style transfer \cite{gatys2016image}. Features were taken from all outputs of FPN layers. Therefore, in addition to $B$ RoIs from which we compute $L^S$, we add the loss from features extracted from the whole image. The objective of adding this loss is to improve the Generator's ability to output a more neutral, e.g. black, background.\\
    \item Full model: we combine base model and all three extensions
\end{itemize}
We trained in total five frameworks (baseline + three augmentations + full model). Each framework was trained for 500 epochs, using Adam optimizer($\beta_1 = 0.9, \beta_2 = 0.999$), regularization parameter (weight decay) of $1e-3$. Hyperparameters of Faster R-CNN logo detector were the same across all frameworks, and shared most of them with the pre-trained logo detector, including the size of the RoIs, $H=7, W=7, C=256$. Since logo generation is a very spatially sensitive task, we used different thresholds for positive and negative candidates both at RPN and RoIAlign stages: the positive threshold was $0.9$ and negative $0.1$.   
\subsection{StyleGAN2} 
StyleGAN\cite{karras2019style} and StyleGAN2\cite{karras2020analyzing} are the state-of-the art GANs that can learn different styles and generate high-quality large images, this includes training on small dataset ($<$5000 images). We trained StyleGAN2 on our data to generate images size 256$\times$256, using high truncation $\psi=1$ coefficient(no gradient averaging), augment the data by $25\%$, with the learning rate of $1e-4$ for both Generator and Discriminator, Adam optimizer ($\beta_1=0.5, \beta_2=0.999)$, self-attention mechanism \cite{zhang2019self} and batch size of $4$ (maximum possible for this image size on the GPU with 8Gb of VRAM. We trained each model (with and without attention modules) for 100000 steps ($\sim$ 100 epochs), which took about 72 hours, but we noticed that after about 20000 steps the model starts to overfit and exhibits a strong mode collapse. We therefore report the best result for each model (20000 steps for the StyleGAN2 with attention and 15000 for StyleGAN2 without attention).  
\subsection{Self-Attention GANs}
We also train SAGAN, \cite{zhang2019self}, with spectral normalization\cite{miyato2018spectral} and Hinge loss function. We used the recommended hyperparameters: latent dimension size $128$, batch size of $64$, Generator learning rate $1e-4$, Discriminator learning rate $4e-4$ and Adam optimizer ($\beta_1 = 0, \beta_2=0.9$). Generator's architecture consists of 7 modules ($ConvTranspose2D + BatchNorm+ReLU$, each equipped with a spectral transformer. Self-attention module is added to block $3$ with $256$ feature maps and map size of $16 \times 16$. The model outputs images size 256$\times$256. SAGAN framework was trained for $300000$ iterations ($\sim 330$ epochs). Training was stopped due to the obvious mode collapse. 

\section{Evaluation of Results}
\label{sec:accuracy}
Examples of outputs of all models are presented in Figure \ref{fig:output_generators}. In Table \ref{tab:accuracy_results_fid_is} we report FID and IS scores, in Table \ref{tab:accuracy_results} we report quality and detection results for all models. The best results are bold+italicized, second best bold and third-best italicized. For FID score, we used the layer with $2048$ maps, for IS scores we split the sample into either $1$ or $10$ subsets. Each model generates 512 images which are processed by Faster R-CNN logo detector. If it predicts a logo with confidence score exceeding the pre-defined threshold of 0.75, the detection is considered to be a True Positive (TP), otherwise it is a False Positive (FP). The assumption of this test is that a good Generator would output images that contain exactly single identifiable logo. If the detector predicts more than one logo in a single image with confidence exceeding this threshold, all predictions other than the best-scored one are counted as FPs. If it predicts no logos at all, it is also counted as an FP. Detection rate is defined as $\frac{TP}{TP+FP}$, average confidence is averaged over all detections, including those below the threshold.
\subsection{DCGAN+ and LL-GAN}
\begin{figure*}
    \centering
    \includegraphics[width=0.35\linewidth]{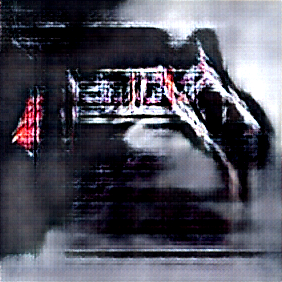}
    \vline \includegraphics[width=0.35\linewidth]{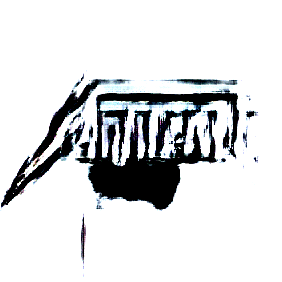}\\
    \includegraphics[width=0.35\linewidth]{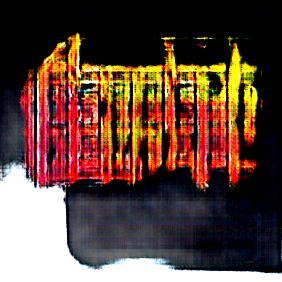}
    \vline \includegraphics[width=0.35\linewidth]{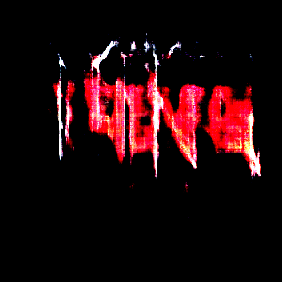}\\
        \includegraphics[width=0.35\linewidth]{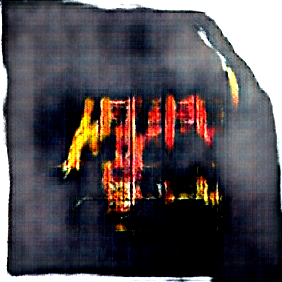}
    \vline \includegraphics[width=0.35\linewidth]{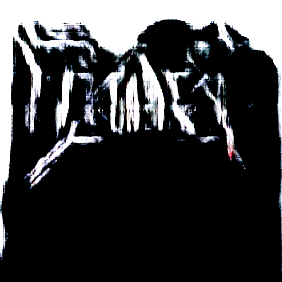}\\
    \includegraphics[width=0.35\linewidth]{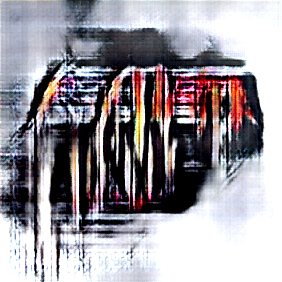}
    \vline \includegraphics[width=0.35\linewidth]{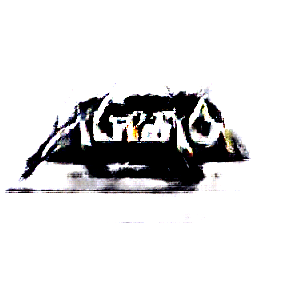}
    \caption{Comparison of DCGAN+ (left) and LL-GAN output (right). First row: DCGAN+ vs LL-GAN, second row: DCGAN+ vs LL-GAN(+backbone features), third row: DCGAN+ vs LL-GAN (full), fourth row: DCGAN+ vs LL-GAN(+FRCNN losses). The obvious weakness of DCGAN+ that LL-GAN fixes is the lack of shape (glyphs are made up of thicker, shorter features without gaps) and color (all glyphs in the logo have the same color) consistency. Each row used the same Generator input. Best viewed in color.}
    \label{fig:dcgan_plus_vs_llgan}
\end{figure*}

DCGAN+ achieves the best FID score of $220.155$, in which it confidently outperforms far more sophisticated state-of-the-art models. It also achieves the third-best results across all other scores. The baseline model is capable of producing high-quality realistic logos in the style of heavy metal bands without overfitting to any particular feature. Among its weaknesses are the inconsistency in glyph stlye, both in terms of color and background noise, see Figures \ref{fig:dcgan_plus_vs_llgan} and \ref{fig:output_generators}. In particular, some logos are red and yellow and consist of thin vertical lines. Vanilla LL-GAN model achieves the best IS scores of $6.339$ and $5.292$ and outputs highly detectable logos with high confidence. 
\begin{table}
  \centering
      \caption{Comparison of models' performance-Quality. Italicized+bold: best, bold: second-best, italicized: third-best}
  \begin{tabular}{llll}
Framework name & FID & IS(1)& IS(10) \\
\hline
    DCGAN+ &\textbf{\textit{220.155}}&\textit{6.023}&\textit{5.105}\\
    \hline
    LL-GAN & \textbf{223.948} &\textit{\textbf{6.339}}&\textit{\textbf{5.292}}\\
        \ \ + FRCNN loss &271.030&5.705&4.947\\
        \ \ + extended boxes &247.181&5.753&4.901\\
        \ \ + backbone features &\textit{237.752}&4.590&4.095\\
        \ \ full & 249.694 & \textbf{6.232} &\textbf{5.150}\\
    \hline
    StyleGAN2 ($\psi=0.6$) &329.026 &2.840 &2.766 \\
    StyleGAN2 ($\psi=1.0$) &354.873 &2.497&2.433\\
     \ \ +attention &328.859&2.356&2.298\\
    SAGAN &283.554&3.581&3.394
  \end{tabular}
      \label{tab:accuracy_results_fid_is}
\end{table}
Most logos generated by the vanilla model are very realistic, resemble real glyphs, are consistent in colors (mostly red and white, as in the training data), and do not experience mode collapse. Also LL-GAN with all three augmentations perform well, producing IS scores of $6.232$ and $5.150$. In Figure \ref{fig:dcgan_plus_vs_llgan} we placed outputs from DCGAN+ and different LL-GAN models that output logos with similar features side-by-side to highlight the advantages of our approach. The same features produced by LL-GAN generators are more homogeneous in color and shape, the background contains fewer geometric artefacts and is more consistent and neutral. Metrics discussed in this section confirm that this consistency does not come at the cost of lower variance in the output.
\subsection{State-of-the-art models}
\begin{figure*}
\centering
 %
\begin{subfigure}{\linewidth}
\centering
\includegraphics[width=0.1\linewidth]{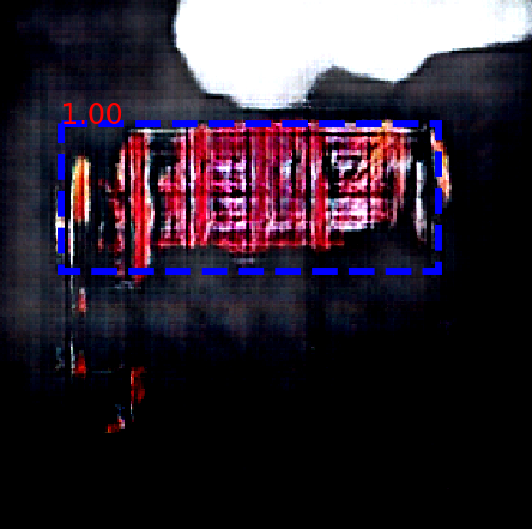}  
\includegraphics[width=0.1\linewidth]{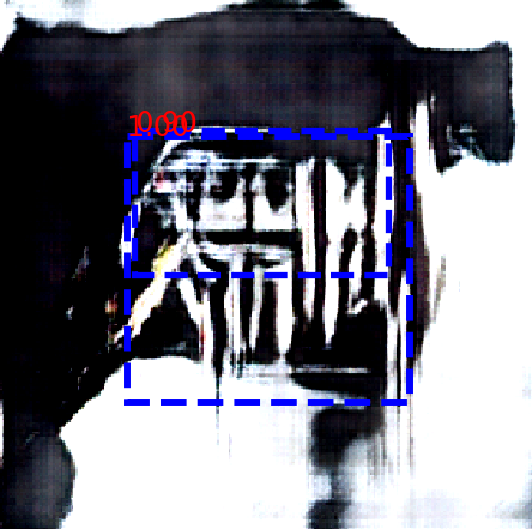}
\includegraphics[width=0.1\linewidth]{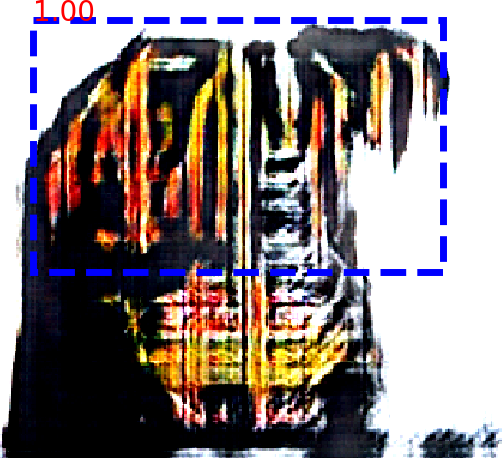} \includegraphics[width=0.1\linewidth]{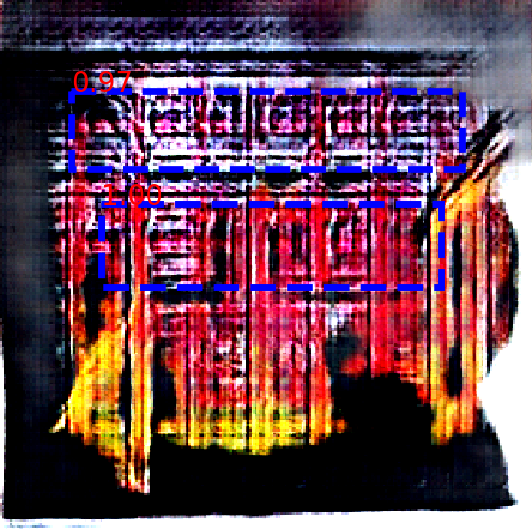}
\includegraphics[width=0.1\linewidth]{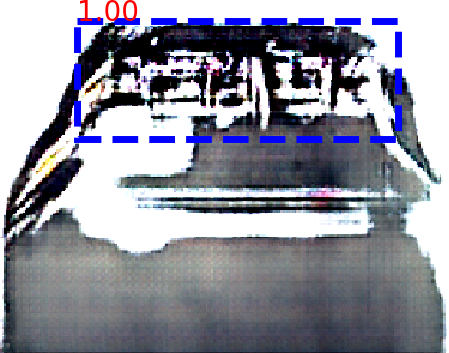} \includegraphics[width=0.1\linewidth]{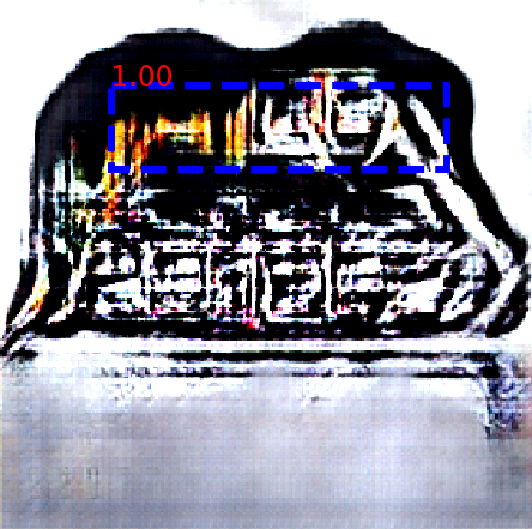}
\includegraphics[width=0.1\linewidth]{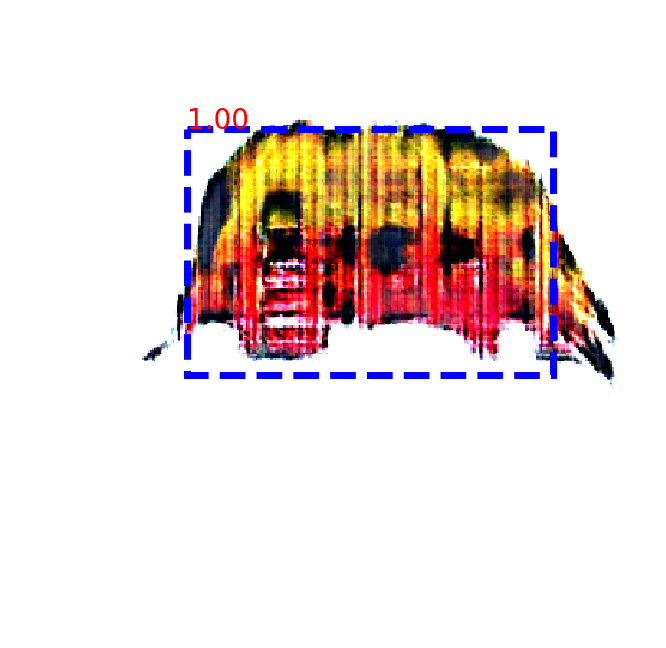}
\includegraphics[width=0.1\linewidth]{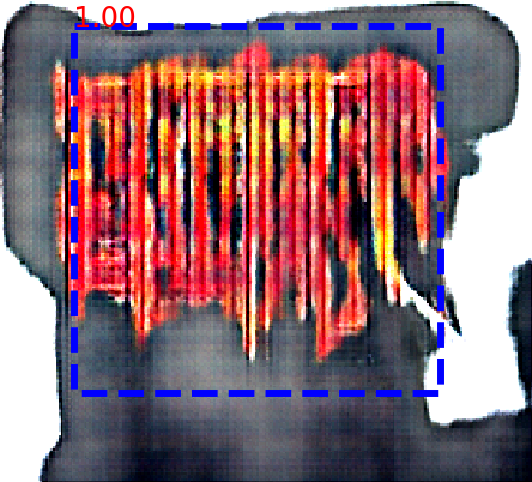}
\includegraphics[width=0.1\linewidth]{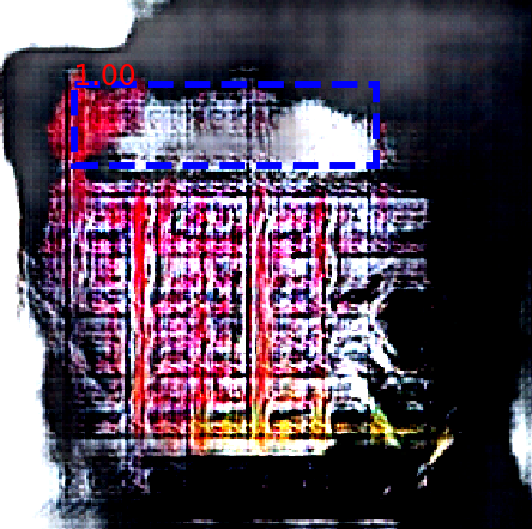}
\subcaption{DCGAN+}
\end{subfigure}
\begin{subfigure}{\linewidth}
  \centering
  \includegraphics[width=0.1\linewidth]{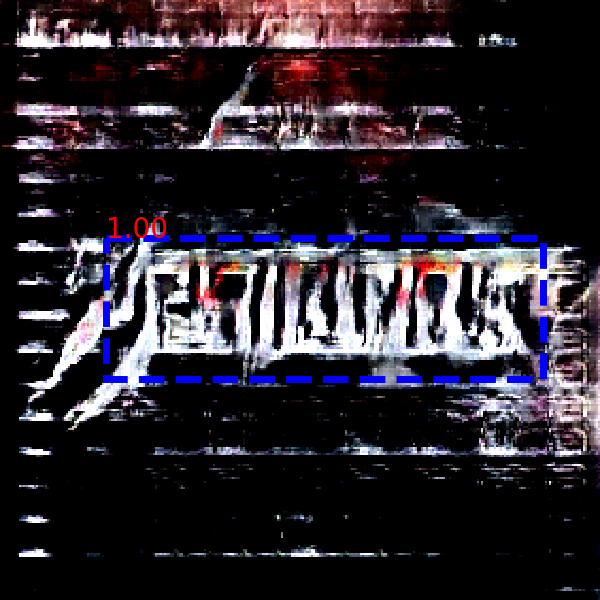}
    \includegraphics[width=0.1\linewidth]{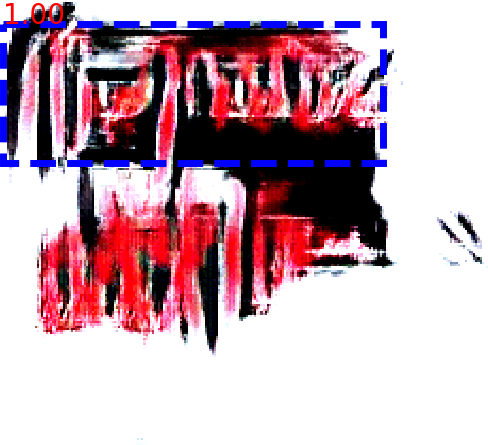}  \includegraphics[width=0.1\linewidth]{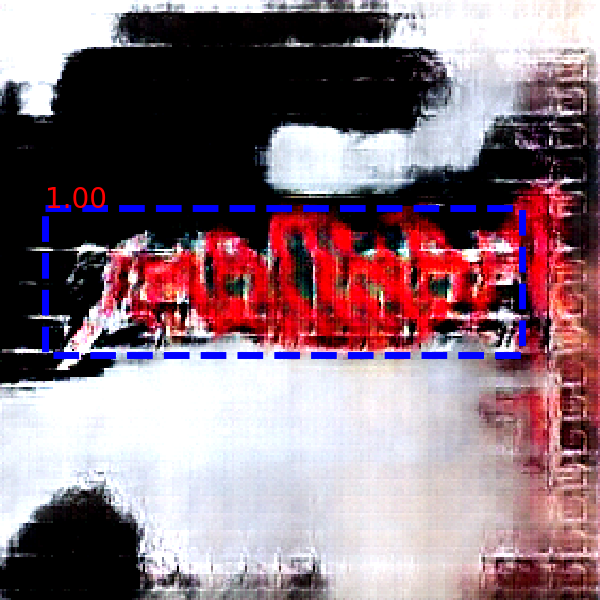}  \includegraphics[width=0.1\linewidth]{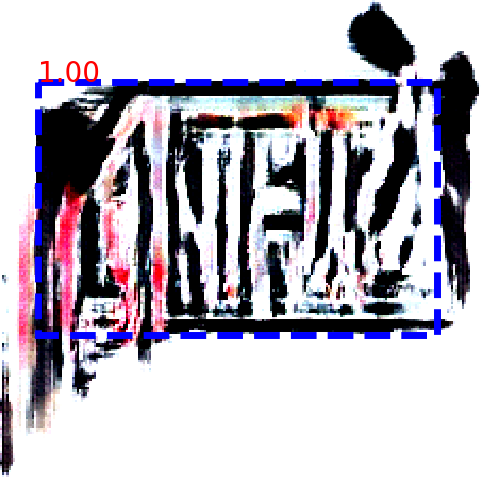}  \includegraphics[width=0.1\linewidth]{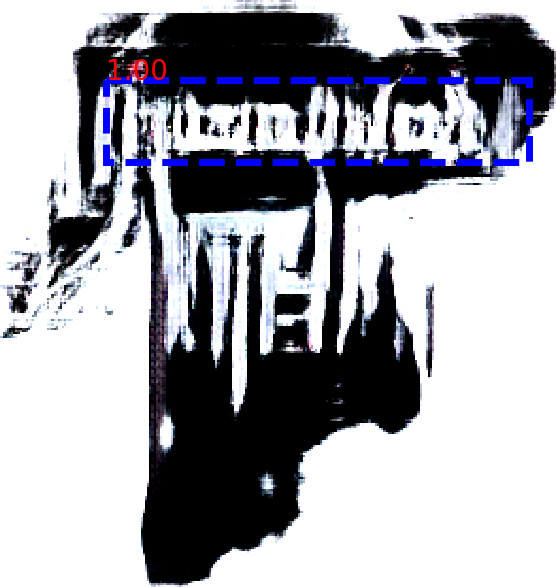}  \includegraphics[width=0.1\linewidth]{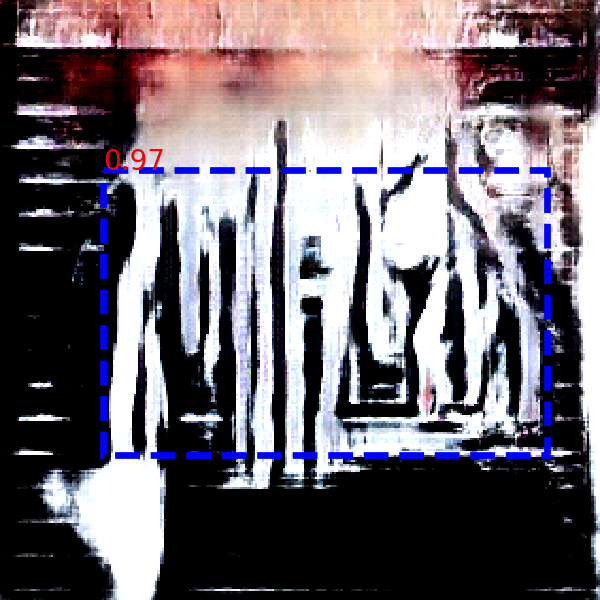}  \includegraphics[width=0.1\linewidth]{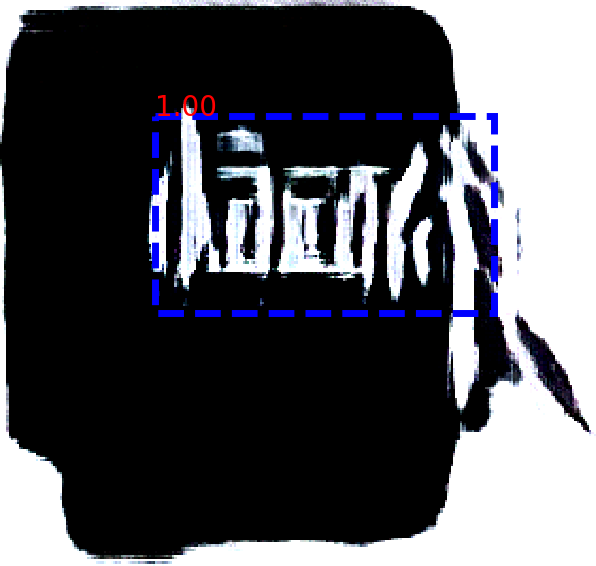}  \includegraphics[width=0.1\linewidth]{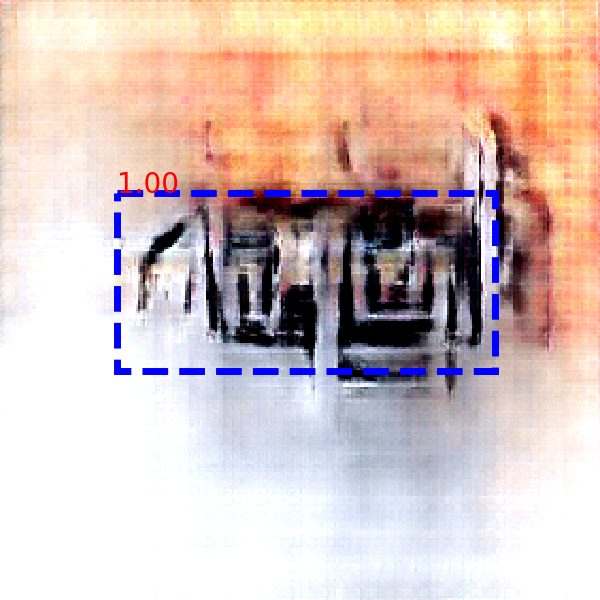}  \includegraphics[width=0.1\linewidth]{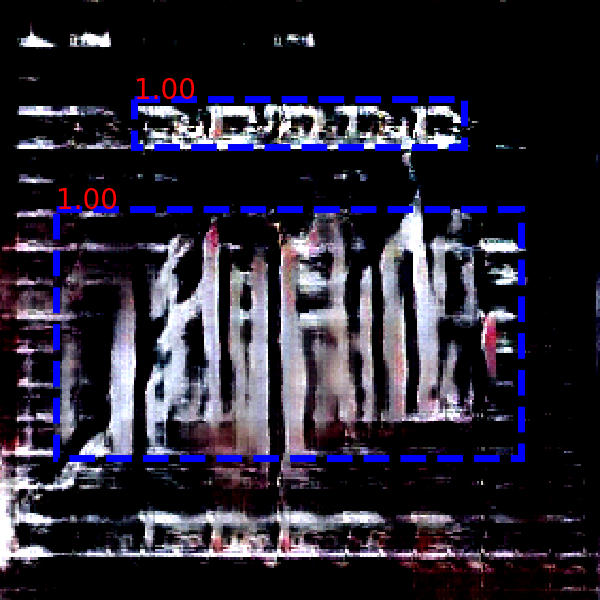}
  \subcaption{LL-GAN}
  \end{subfigure}
  \begin{subfigure}{\linewidth}
  \centering
\includegraphics[width=0.1\linewidth]{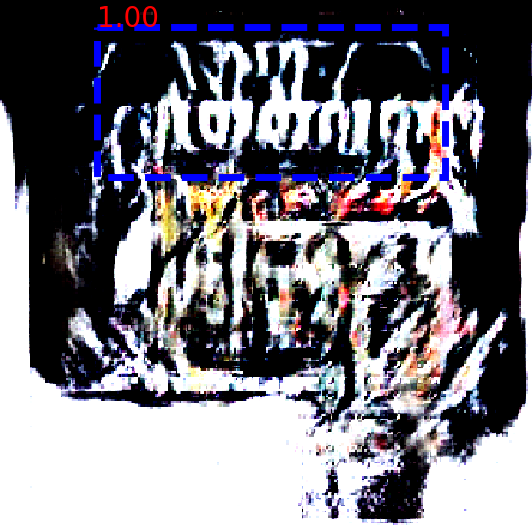}
\includegraphics[width=0.1\linewidth]{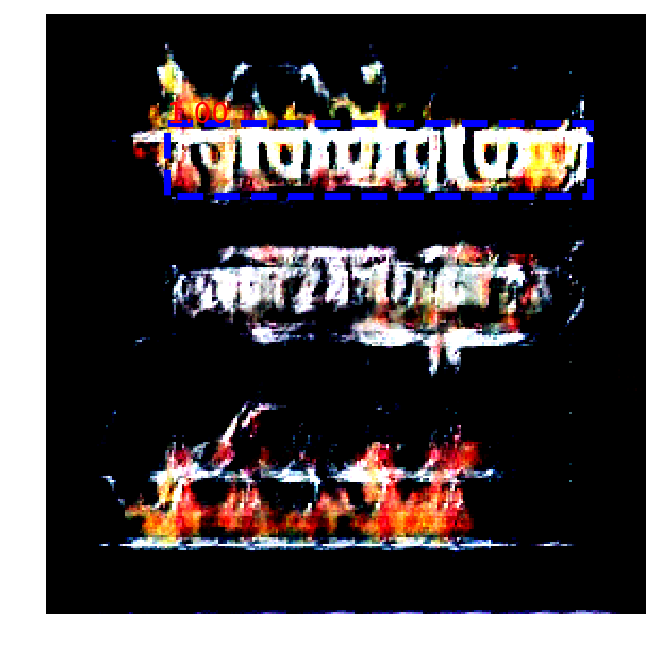}
\includegraphics[width=0.1\linewidth]{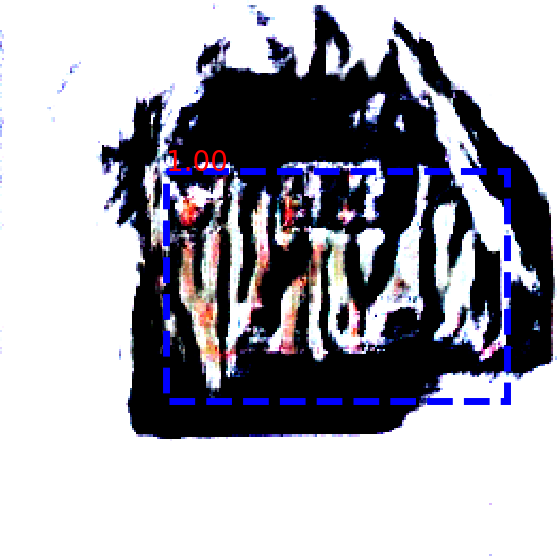}
\includegraphics[width=0.1\linewidth]{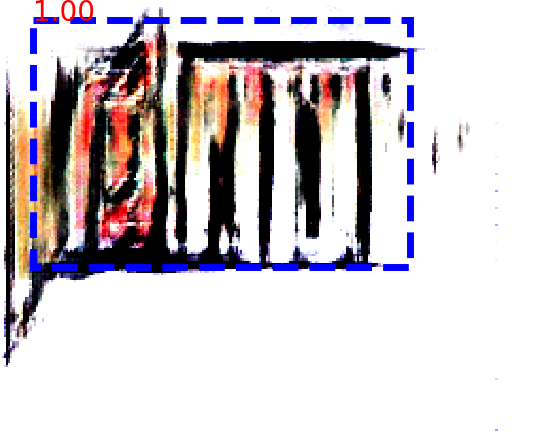}
\includegraphics[width=0.1\linewidth]{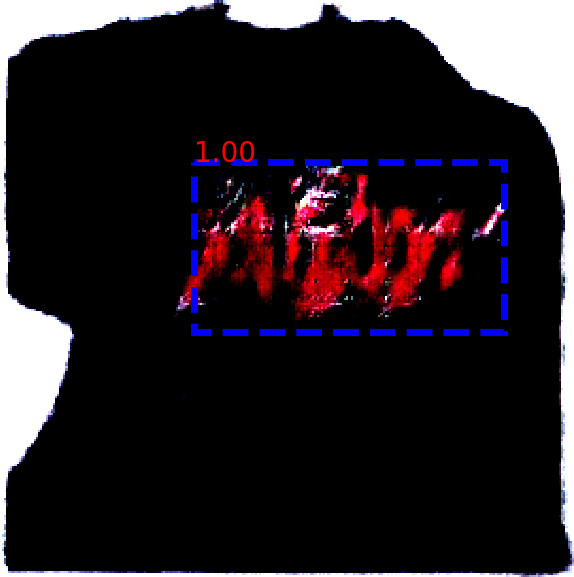}
\includegraphics[width=0.1\linewidth]{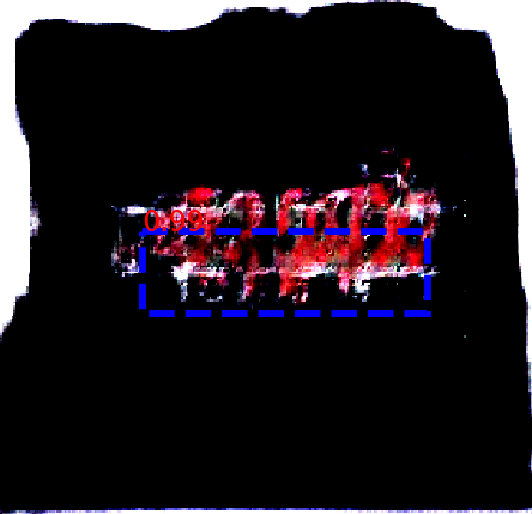}
\includegraphics[width=0.1\linewidth]{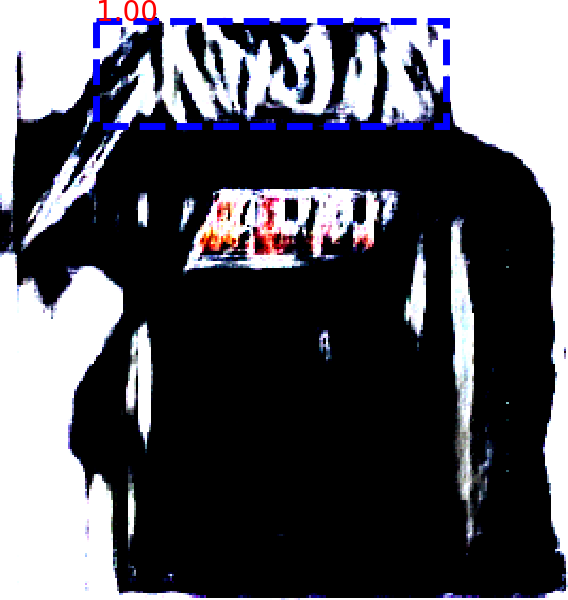}
\includegraphics[width=0.1\linewidth]{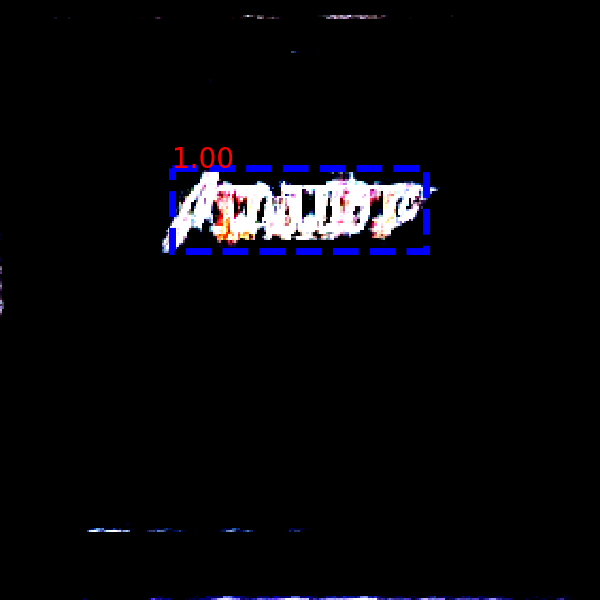}
\includegraphics[width=0.1\linewidth]{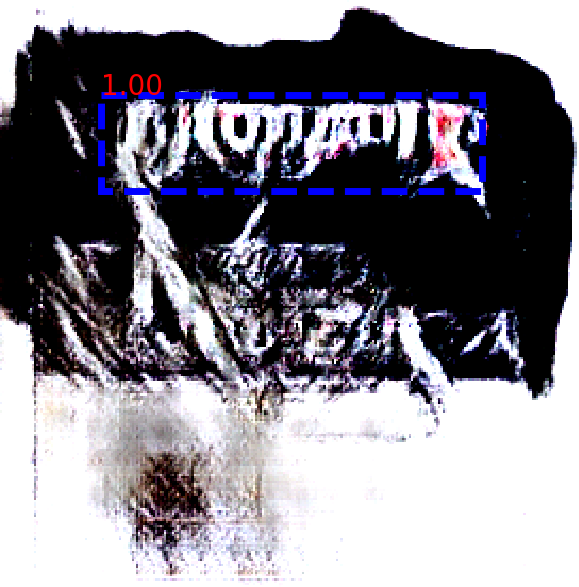}
  \subcaption{LL-GAN + extended boxes}
  \end{subfigure}
    \begin{subfigure}{\linewidth}
  \centering
\includegraphics[width=0.1\linewidth]{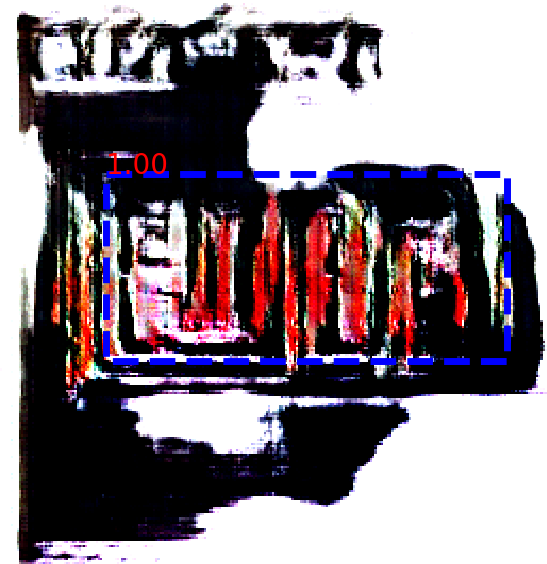}
\includegraphics[width=0.1\linewidth]{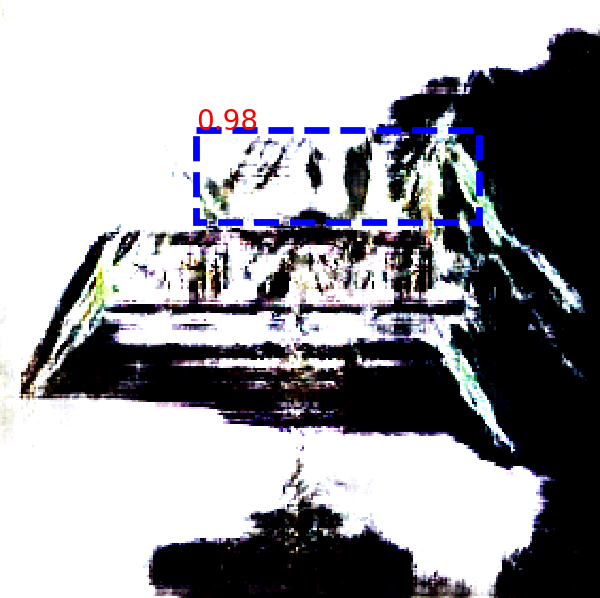}
\includegraphics[width=0.1\linewidth]{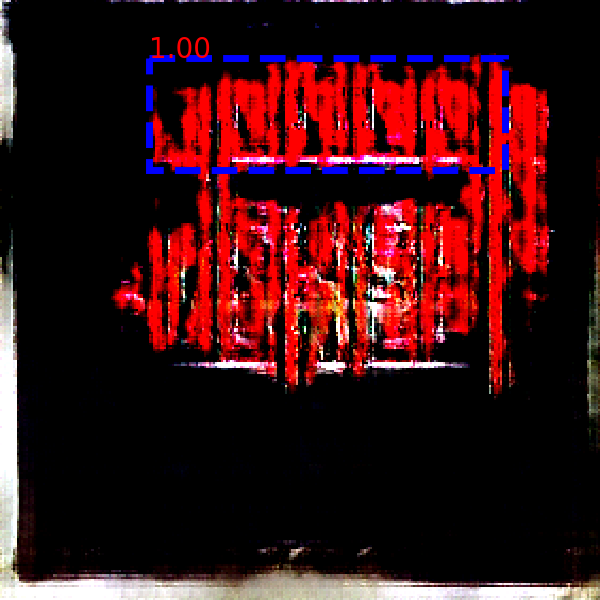}
\includegraphics[width=0.1\linewidth]{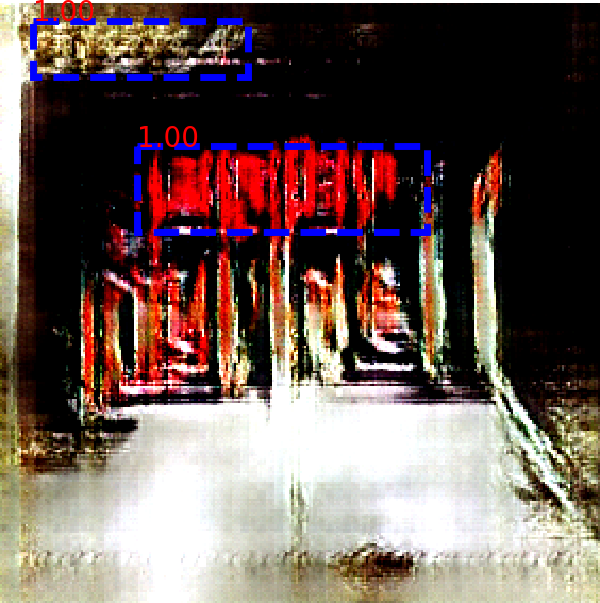}
\includegraphics[width=0.1\linewidth]{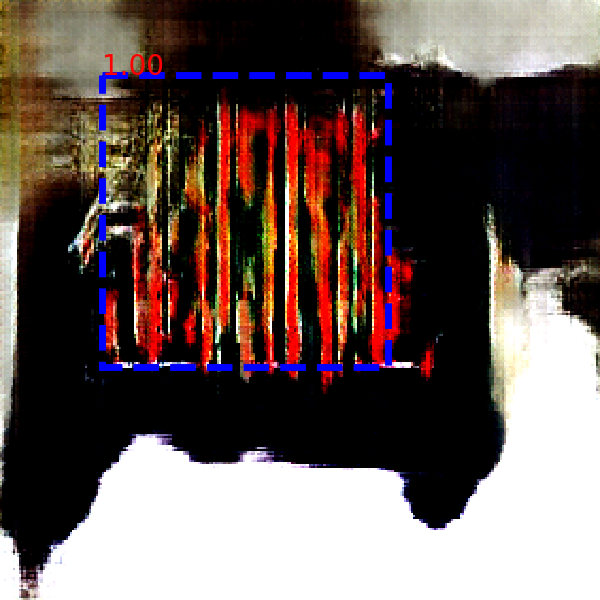}
\includegraphics[width=0.1\linewidth]{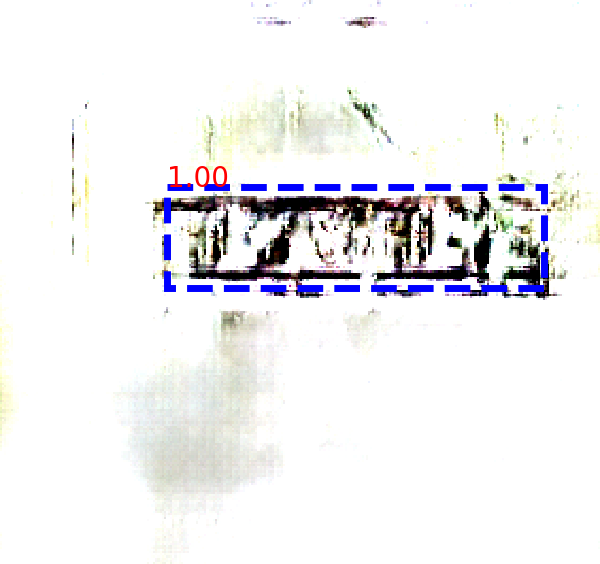}
\includegraphics[width=0.1\linewidth]{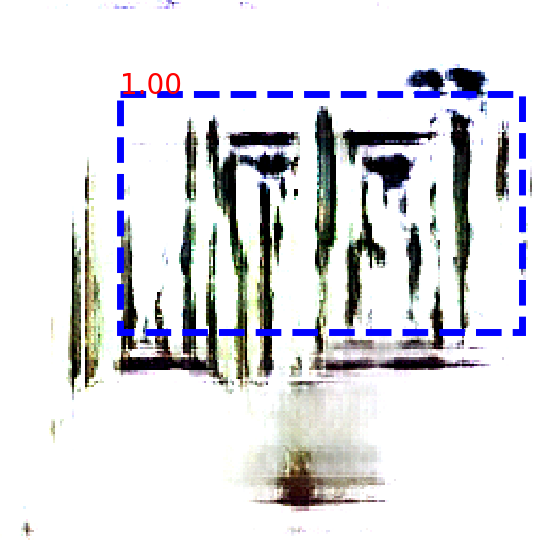}
\includegraphics[width=0.1\linewidth]{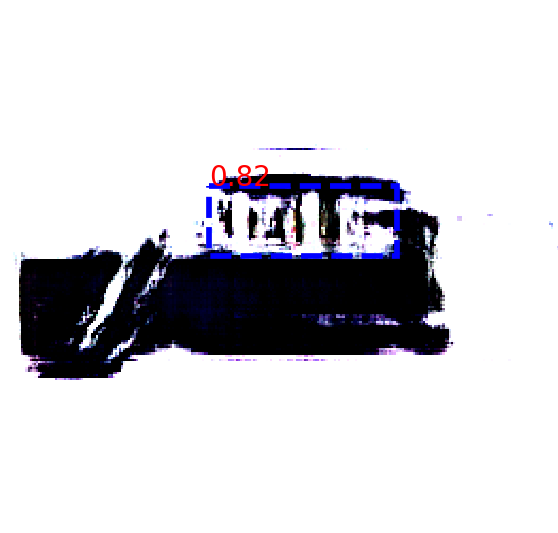}
\includegraphics[width=0.1\linewidth]{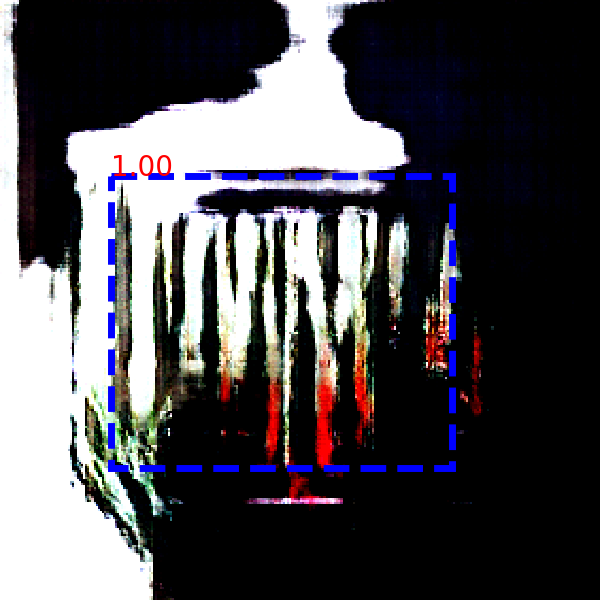}
  \subcaption{LL-GAN + Faster R-CNN loss}
  \end{subfigure}
    \begin{subfigure}{\linewidth}
  \centering
\includegraphics[width=0.1\linewidth]{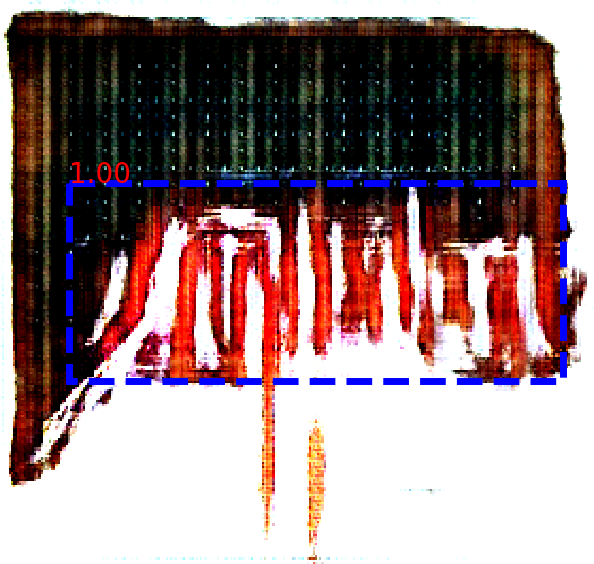}
\includegraphics[width=0.1\linewidth]{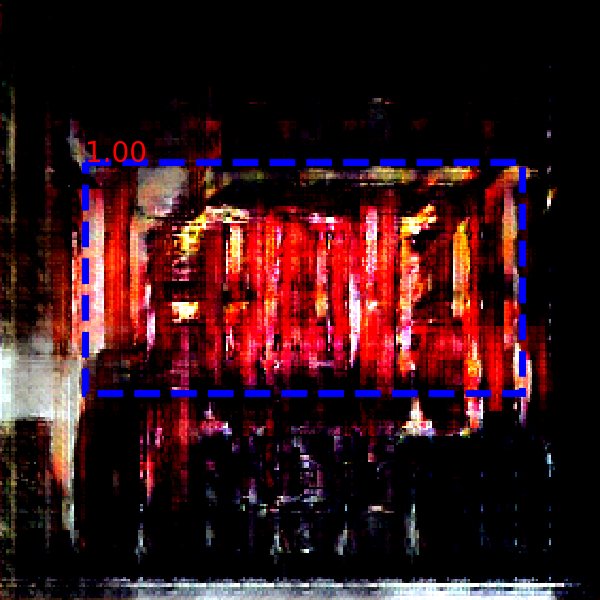}
\includegraphics[width=0.1\linewidth]{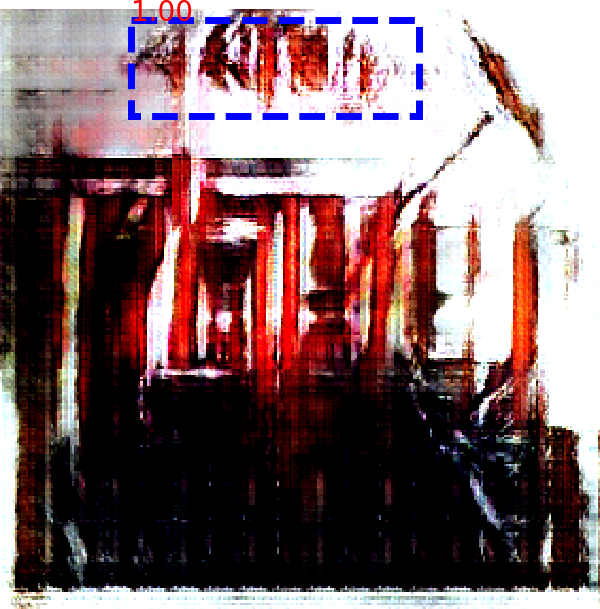}
\includegraphics[width=0.1\linewidth]{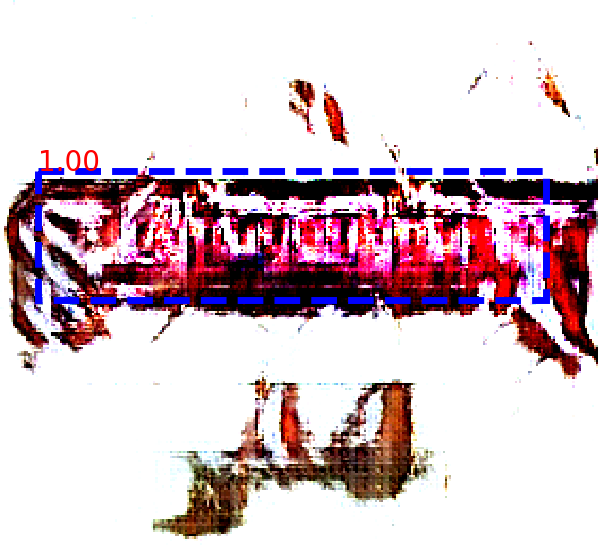}
\includegraphics[width=0.1\linewidth]{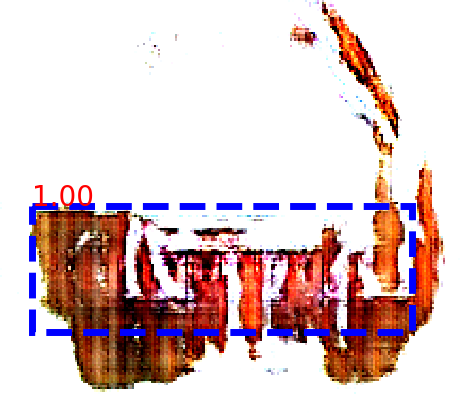}
\includegraphics[width=0.1\linewidth]{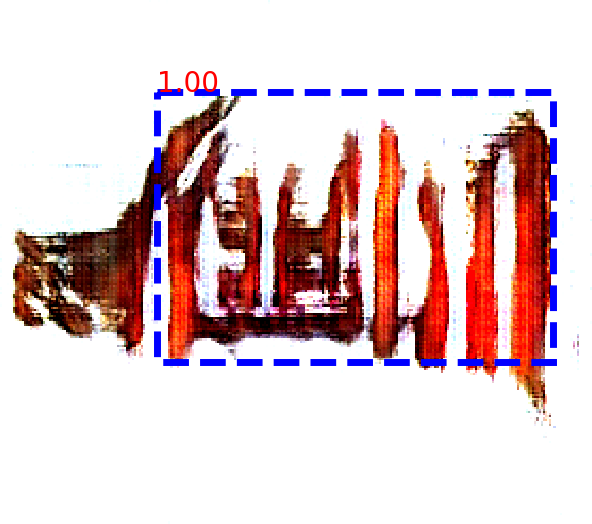}
\includegraphics[width=0.1\linewidth]{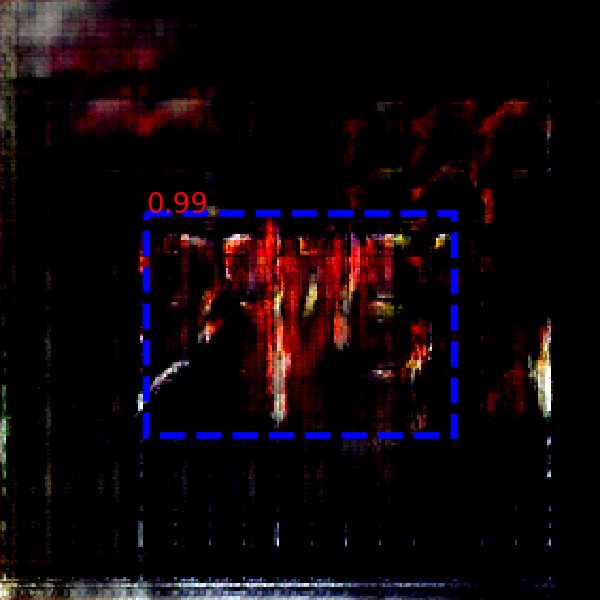}
\includegraphics[width=0.1\linewidth]{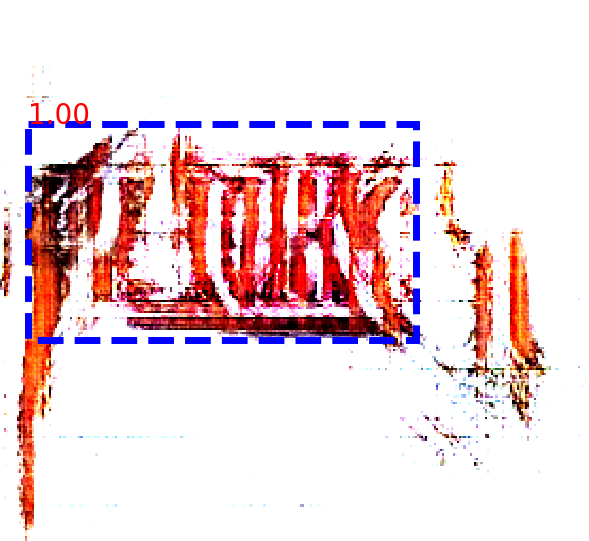}
\includegraphics[width=0.1\linewidth]{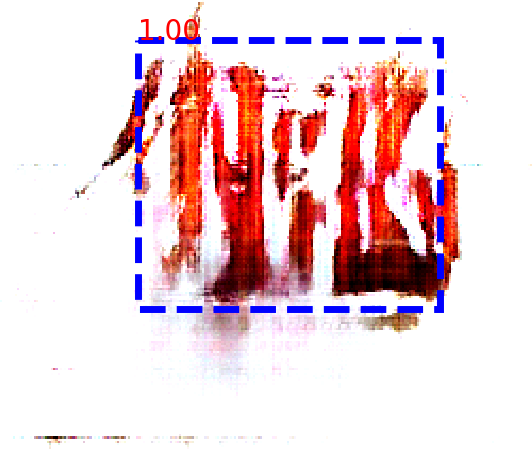}
  \subcaption{LL-GAN + backbone features}
  \end{subfigure}
      \begin{subfigure}{\linewidth}
  \centering
\includegraphics[width=0.1\linewidth]{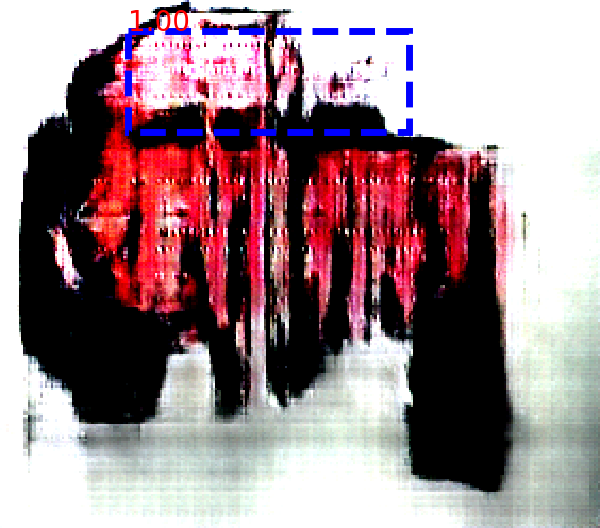}
\includegraphics[width=0.1\linewidth]{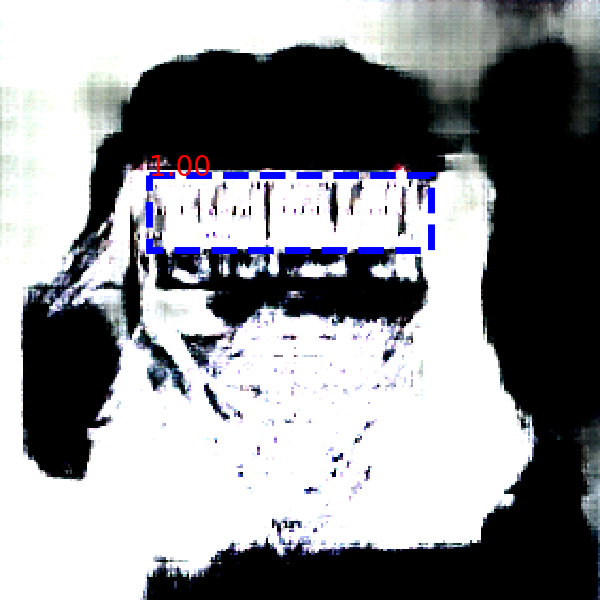}
\includegraphics[width=0.1\linewidth]{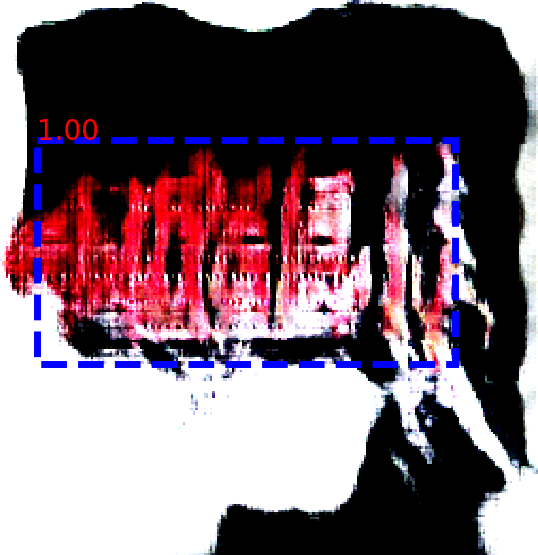}
\includegraphics[width=0.1\linewidth]{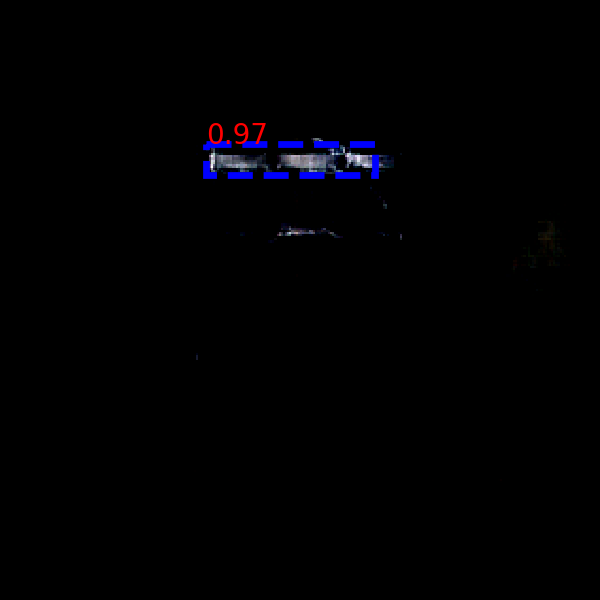}
\includegraphics[width=0.1\linewidth]{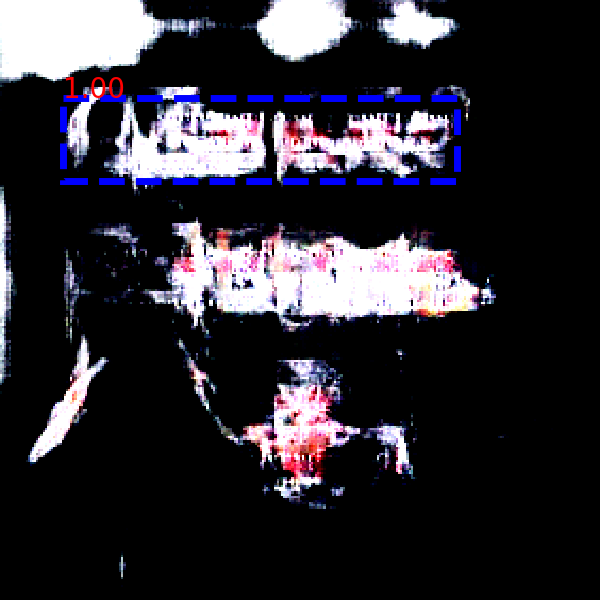}
\includegraphics[width=0.1\linewidth]{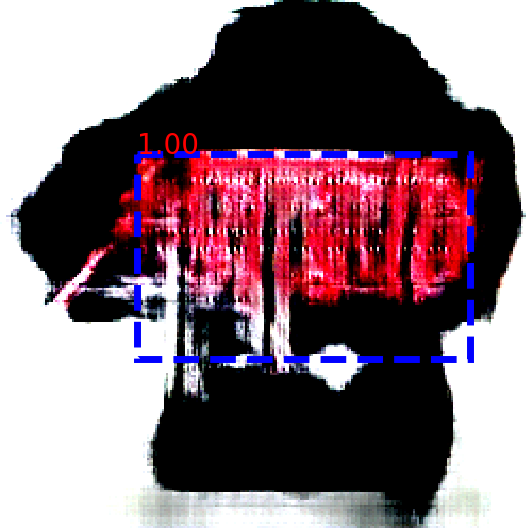}
\includegraphics[width=0.1\linewidth]{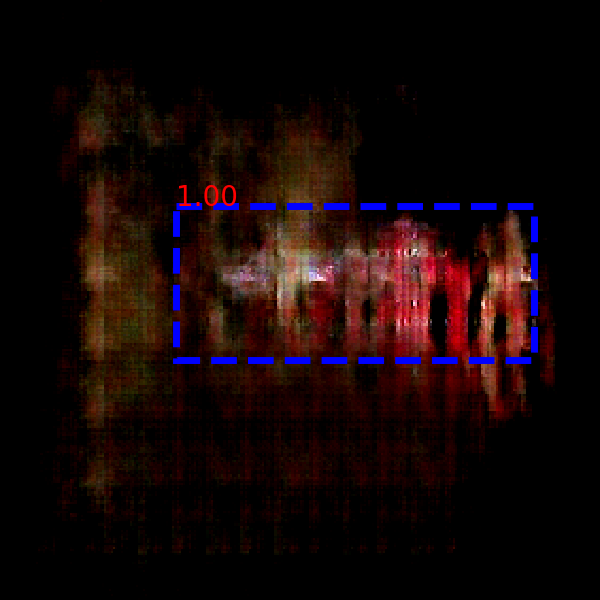}
\includegraphics[width=0.1\linewidth]{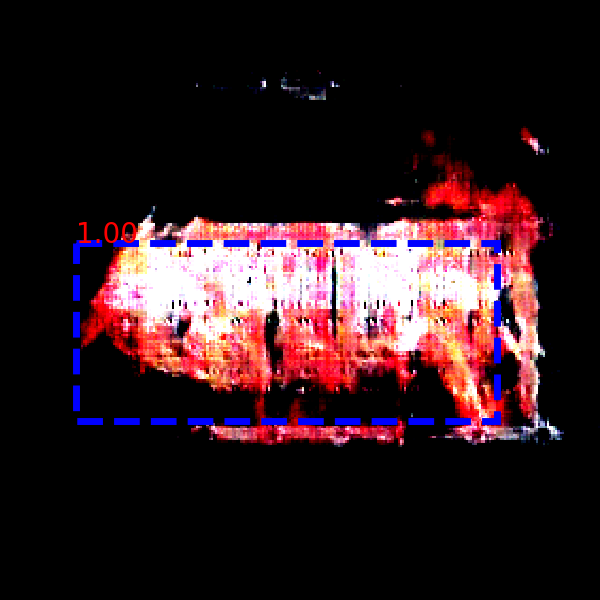}
\includegraphics[width=0.1\linewidth]{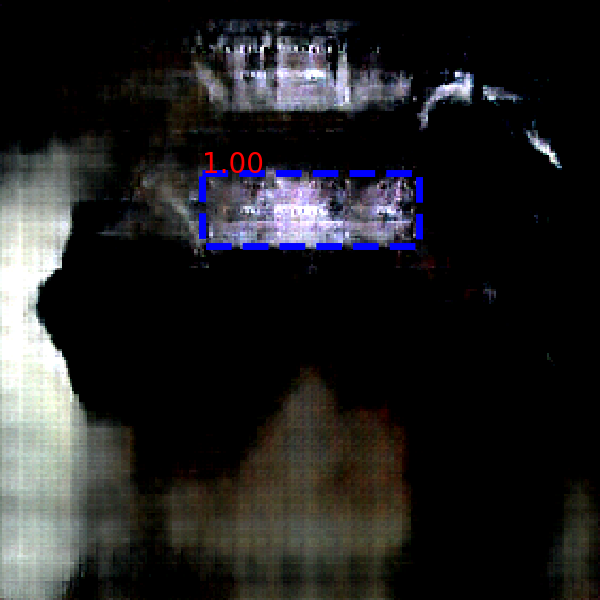}
  \subcaption{LL-GAN (full)}
  \end{subfigure}

  \begin{subfigure}{\linewidth}
  \centering
    \includegraphics[width=0.1\linewidth]{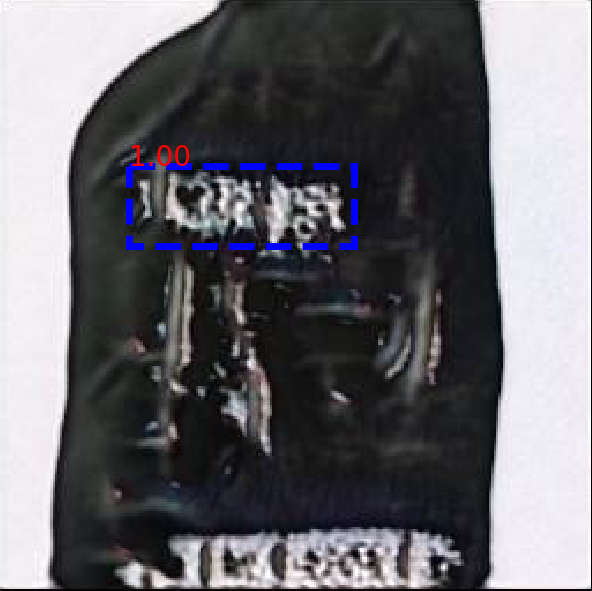}                   \includegraphics[width=0.1\linewidth]{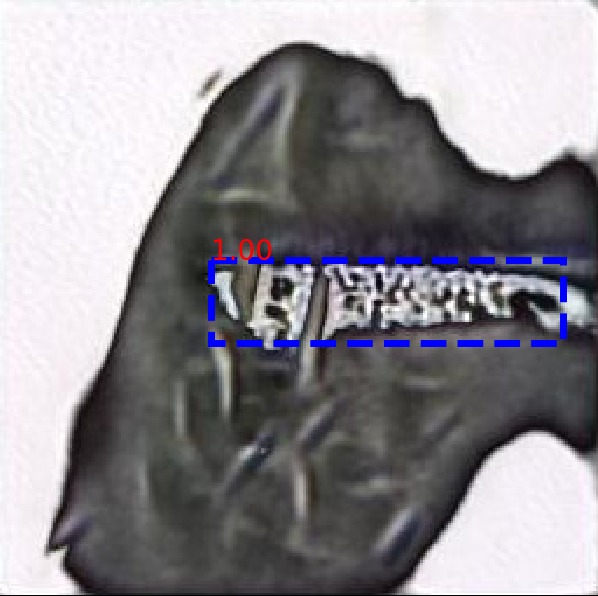}
    \includegraphics[width=0.1\linewidth]{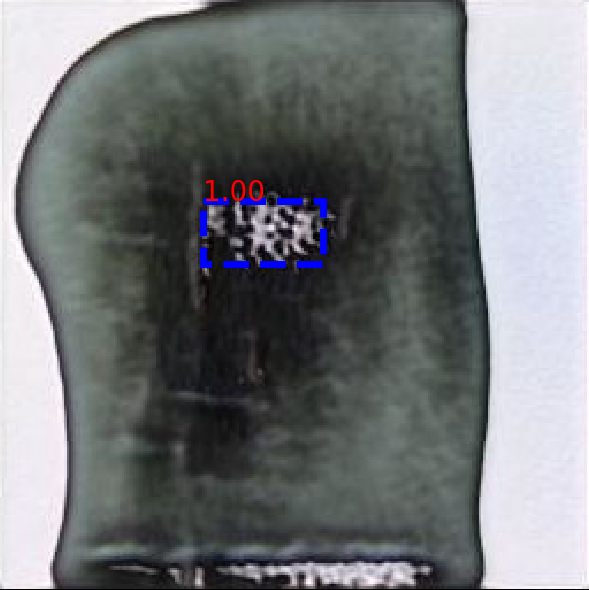}
    \includegraphics[width=0.1\linewidth]{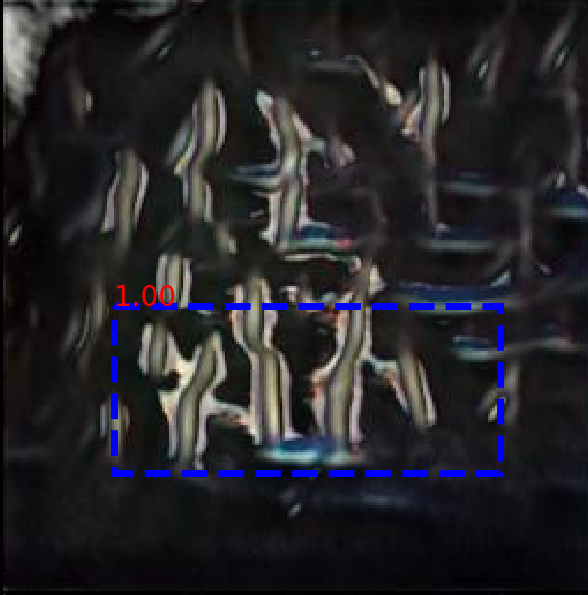}
    \includegraphics[width=0.1\linewidth]{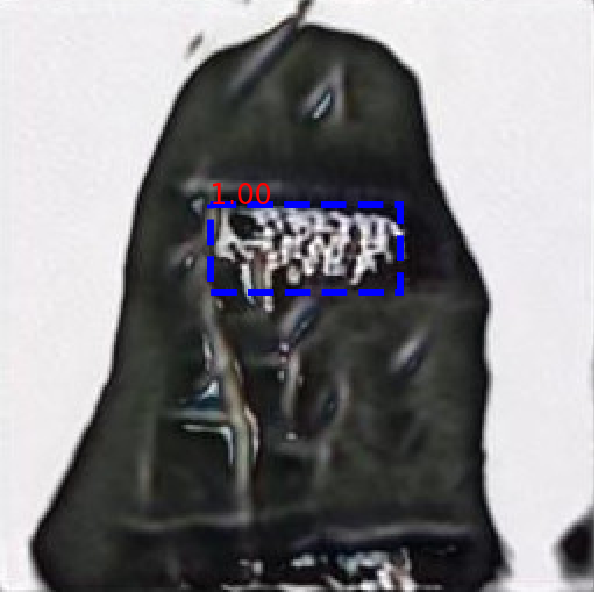}
    \includegraphics[width=0.1\linewidth]{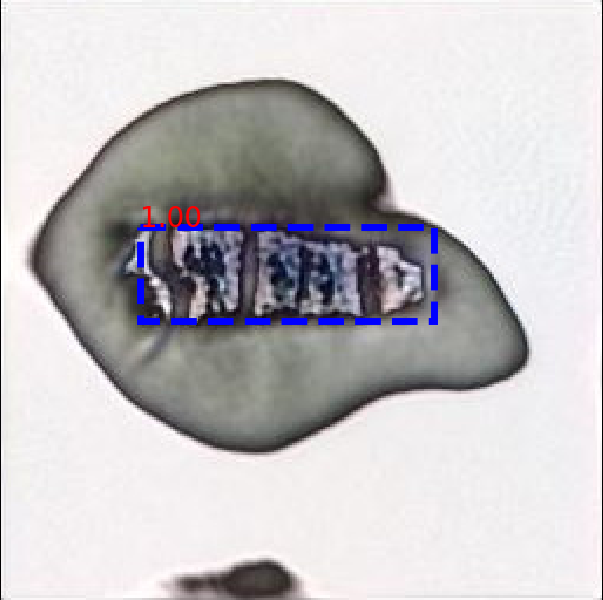}
    \includegraphics[width=0.1\linewidth]{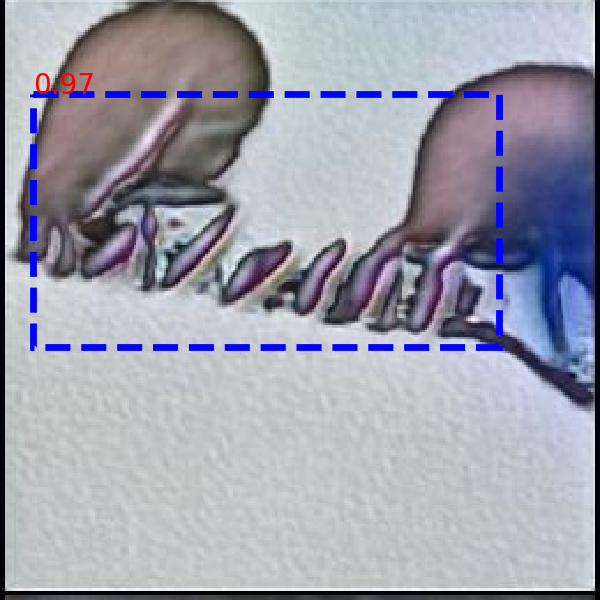}
    \includegraphics[width=0.1\linewidth]{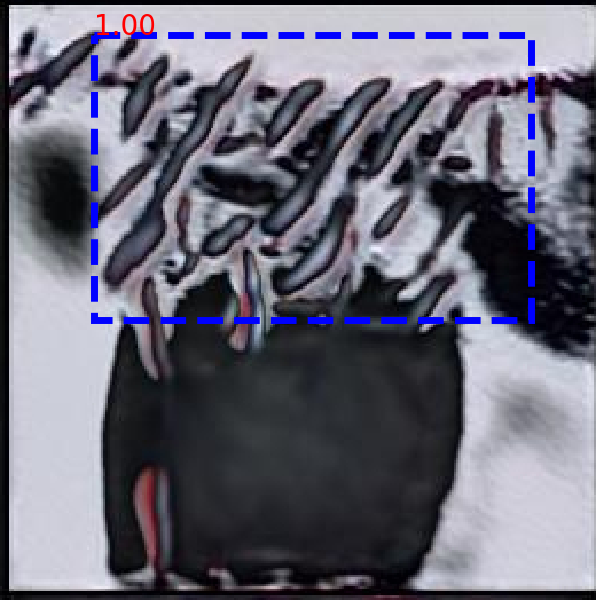}
    \includegraphics[width=0.1\linewidth]{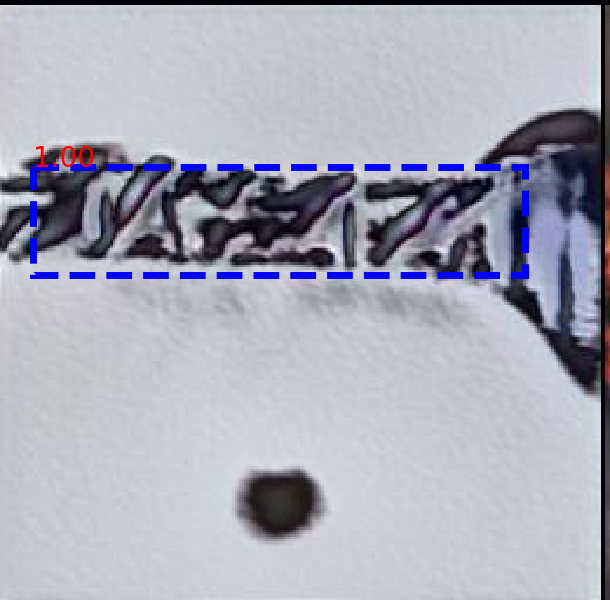}
    \subcaption{StyleGAN2 ($\psi=1$)}
    \end{subfigure}
    \begin{subfigure}{\linewidth}
    \centering
    \includegraphics[width=0.1\linewidth]{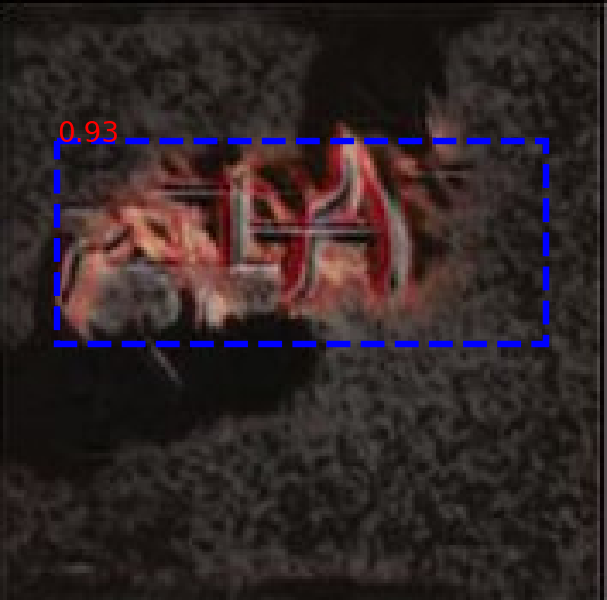}                   \includegraphics[width=0.1\linewidth]{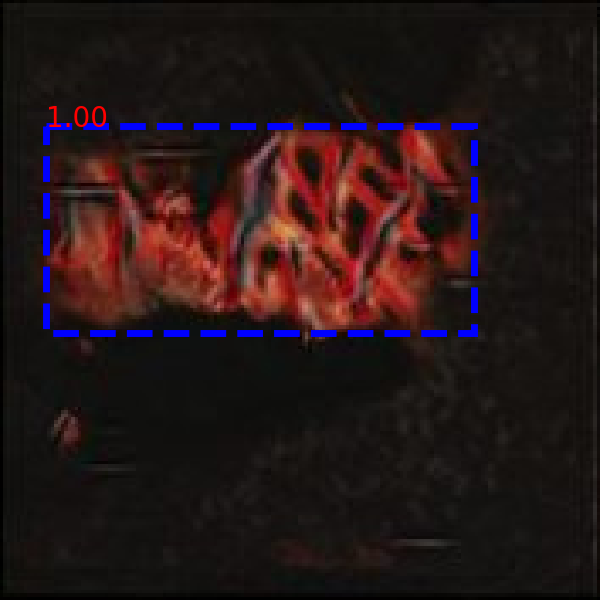}
    \includegraphics[width=0.1\linewidth]{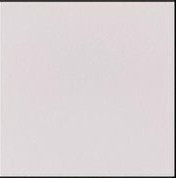}
    \includegraphics[width=0.1\linewidth]{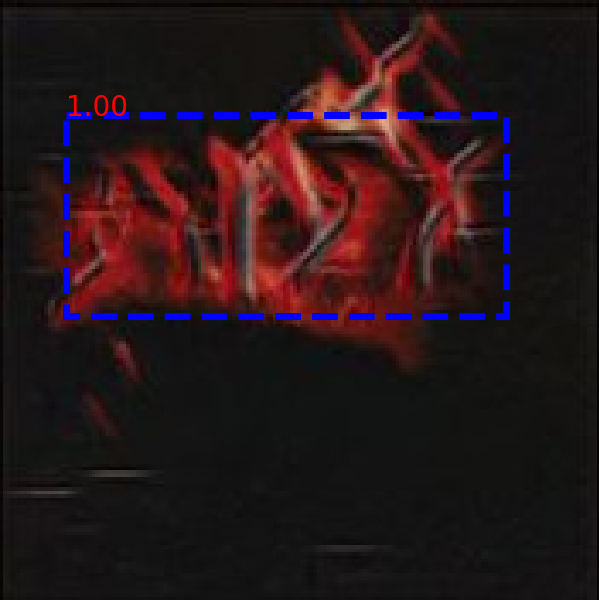}
    \includegraphics[width=0.1\linewidth]{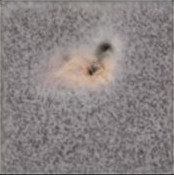}
    \includegraphics[width=0.1\linewidth]{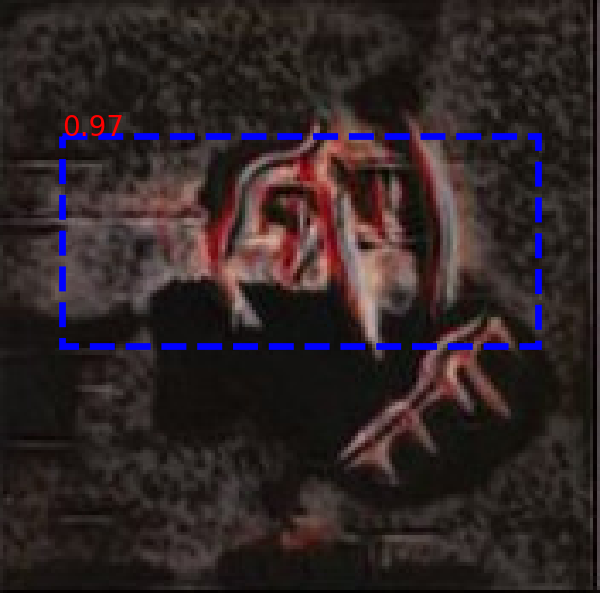}
        \includegraphics[width=0.1\linewidth]{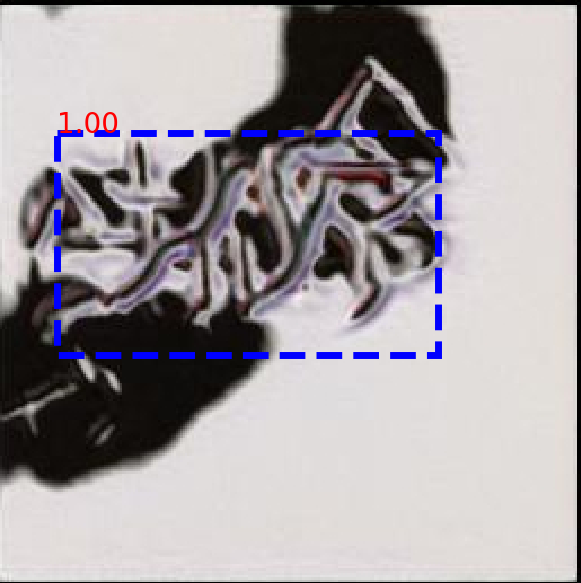}
            \includegraphics[width=0.1\linewidth]{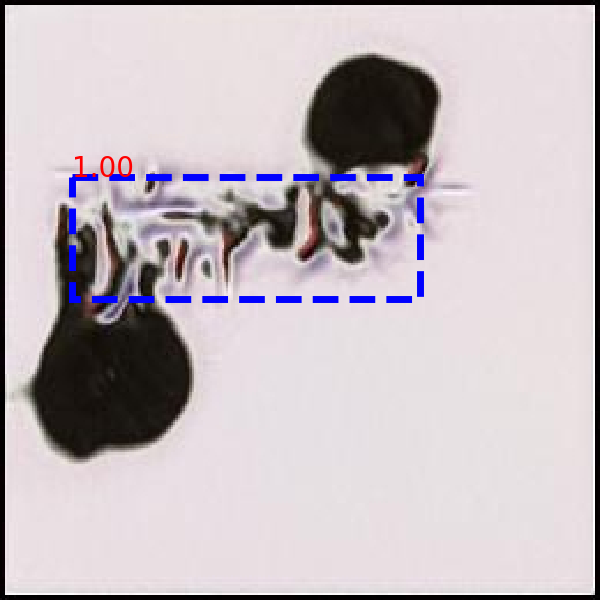}
                \includegraphics[width=0.1\linewidth]{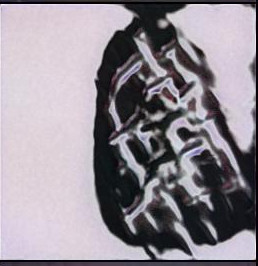}
    \subcaption{StyleGAN2 ($\psi=1$) + attention module}
    \end{subfigure}
    \begin{subfigure}{\linewidth}
    \centering
        \includegraphics[width=0.1\linewidth]{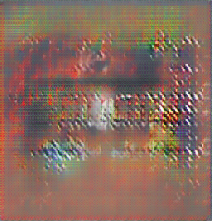}                   
    \includegraphics[width=0.1\linewidth]{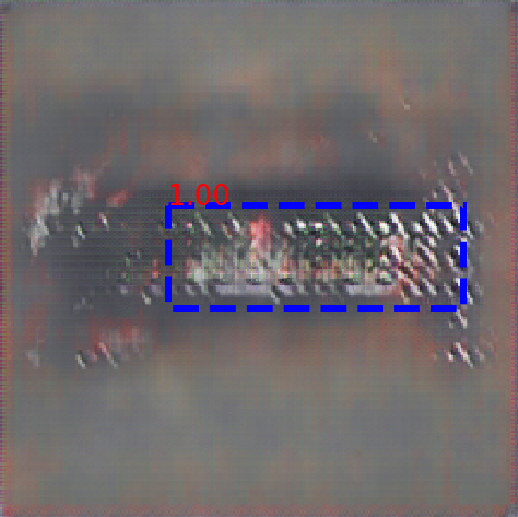}                   \includegraphics[width=0.1\linewidth]{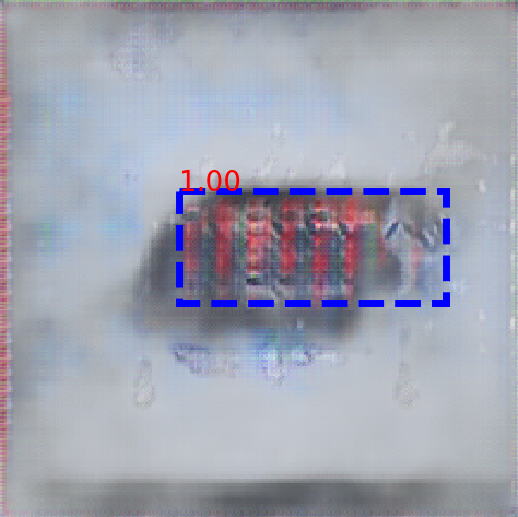}
    \includegraphics[width=0.1\linewidth]{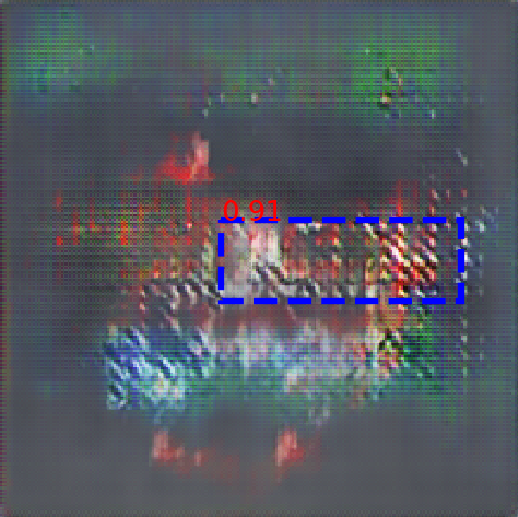}
    \includegraphics[width=0.1\linewidth]{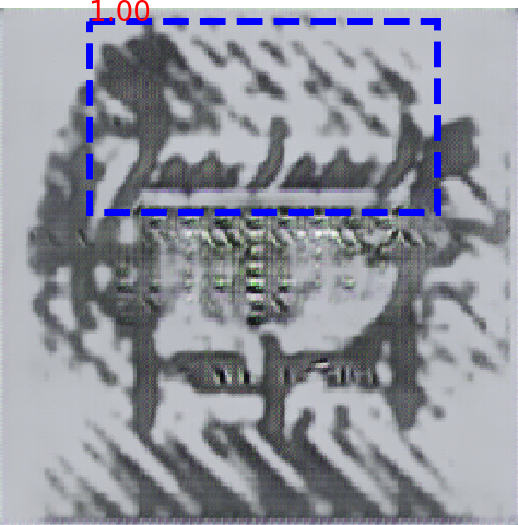}
    \includegraphics[width=0.1\linewidth]{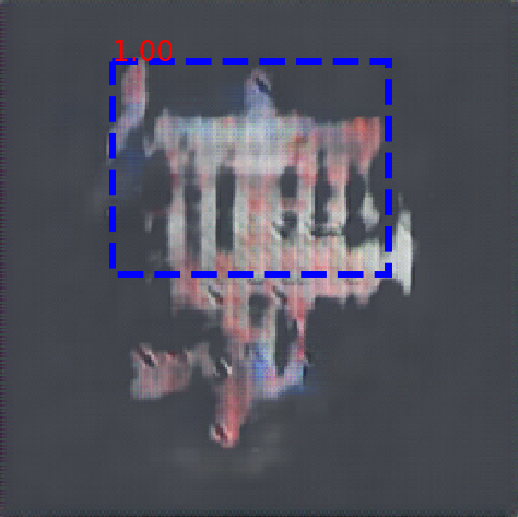}
    \includegraphics[width=0.1\linewidth]{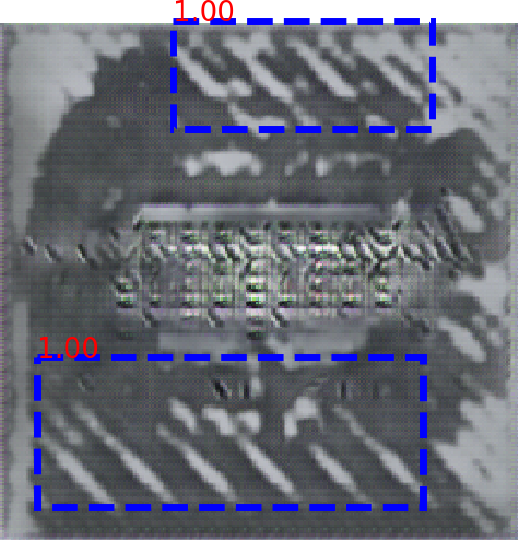}
        \includegraphics[width=0.1\linewidth]{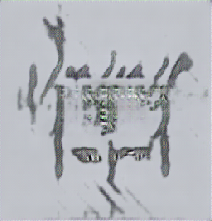}
            \includegraphics[width=0.1\linewidth]{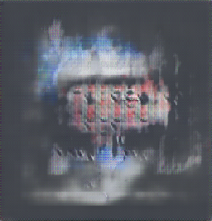}\\
        \subcaption{SAGAN}
    \end{subfigure}
   \caption{Examples generated by the models presented in the paper overlaid with bounding boxes  predicted by the Faster R-CNN logo detection (+confidence score). Three last images for StyleGAN2 and StyleGAN2+Attention models were obtained using mixing regularities, see \cite{karras2020analyzing} for details. All DCGAN+ and LL-GAN images are 282$\times$282, all other models are 256$\times$256. Best viewed in color.}
   \label{fig:output_generators}
\end{figure*}

StyleGAN2 is capable of producing logos with very consistent structures, but due to the size of the dataset suffers from mode collapse. This is reflected in the highest detection score of $0.687$ and low FID and IS scores: the generated structures are consistent enough to be classified as a logo, but do not resemble the training data and are very similar. SAGAN also suffers from mode collapse.\\
\linebreak
By comparing results in Tables \ref{tab:accuracy_results_fid_is} and \ref{tab:accuracy_results} and Figure \ref{fig:output_generators} to the models' architectures and sizes in Table \ref{tab:model_sizes}, LL-GAN models are comparable in size to StyleGAN2, but their Generators output more interesting logos.
\begin{table}
  \centering
  \caption{Comparison of models' performance-Detection. Italicized+bold: best, bold: second-best, italicized: third-best}
  \begin{tabular}{lll}
Framework name & Detection Rate & AvgConf \\
\hline
    DCGAN+ &\textit{0.670}    &  \textit{0.739}\\
    \hline
    LL-GAN &    \textbf{0.674}&  \textbf{0.746}\\
       \ \  + FRCNN loss & 0.640& \textit{\textbf{0.827}}\\
       \ \ + extended boxes&0.666&0.707\\
       \ \ + backbone features &0.622&0.701\\
       \ \ full &0.590 & 0.638\\
       \hline
    StyleGAN2($\psi=0.6$) &0.554&0.670\\
    StyleGAN2($\psi=1.0$) &\textbf{\textit{0.687}}&0.684\\
     \ \ +attention &0.578&0.569\\
    SAGAN & 0.561&0.600
  \end{tabular}
    \label{tab:accuracy_results}
    \end{table}
    
\vspace{-20pt}
\section{Conclusion}
\label{sec:final}
Generation of logos is a challenging problem that is becoming increasingly more popular in deep learning community. In this paper we presented a novel framework that fuses Faster R-CNN and GANs for generating large (282x282) heavy metal logos. The model was trained on a small style-rich dataset of real-life band logos. Results achieved by LL-GAN confidently outperform the state-of-the-art models trained on the same dataset, and we intend to explore the capacity of Faster R-CNN detector to extract and learn from regional features further. The advantages of our approach include:
\begin{itemize}
\item The novel idea of training the Generator using losses extracted from regional features in the real and fake data using Faster R-CNN. \\
\item Computation of the style loss (Gram matrix) on regional features. This allows to use correlation between features in the fake and real data to transfer style from real to fake data, and construct samples from every image.\\ 
\item The use of bounding boxes to determine the size of the RoIs in the fake data. Changing this size can improve results, e.g. by creating a more stable background.
\end{itemize}
Also, we would like to address certain limitations of the presented solution:
\begin{itemize}
    \item Dataset and scope. All models were trained on a small dataset collected specifically to create logos in a particular style. We are confident this approach can be scaled to more general problems (e.g. logo stylization, style transfer, conditional logo creation) and larger datasets.\\   
    \item Disentaglement and fusion of style and content. Disentanglement of style from content is active area of research in the font generation community\cite{gao2019artistic,azadi2018multi}. In this paper we only used a single Generator for the logo generation. This result can be improved both by augmenting the architectures, and fusing the style and content datasets.
\end{itemize}
\bibliographystyle{splncs04}
\bibliography{main_art}
\end{document}